\pdfoutput=1
\documentclass[twocolumn, switch]{article} % Method A for two-column formatting

\usepackage{preprint}

\usepackage{amsmath, amsthm, amssymb, amsfonts}

\usepackage[numbers,square]{natbib}
\usepackage[utf8]{inputenc}    % allow utf-8 input
\usepackage[T1]{fontenc}       % use 8-bit T1 fonts
\usepackage{graphicx}          % required for \includegraphics and \resizebox
\usepackage{multirow}          % required for \multirow in Tables 3, 4, and 6
\usepackage{booktabs}          % professional-quality tables
\usepackage{nicefrac}          % compact symbols for 1/2, etc.
\usepackage{microtype}         % microtypography
\usepackage{lineno}            % Line numbers
\usepackage{float}             % Allows for figures within multicol
\usepackage{lipsum}            % Filler text
\usepackage{xcolor}            % colors for hyperlinks and TikZ
\usepackage{tikz}              % ORCiD icon drawing

\usepackage{newfloat}
\DeclareFloatingEnvironment[name={Supplementary Figure}]{suppfigure}
\usepackage{sidecap}
\sidecaptionvpos{figure}{c}

\usepackage{titlesec}
\titlespacing\section{0pt}{12pt plus 3pt minus 3pt}{1pt plus 1pt minus 1pt}
\titlespacing\subsection{0pt}{10pt plus 3pt minus 3pt}{1pt plus 1pt minus 1pt}
\titlespacing\subsubsection{0pt}{8pt plus 3pt minus 3pt}{1pt plus 1pt minus 1pt}

\usepackage[colorlinks = true,
            linkcolor  = purple,
            urlcolor   = purple,
            citecolor  = orange,
            anchorcolor = black]{hyperref}

\title{Morpho-Spectral Projection Analysis ({MSPA}) for Plant Spectroscopy}

\usepackage{authblk}

\author[1,*]{Lucas Prado Osco}
\author[2]{Marcelo Santos da Silva}
\author[3]{Filipe Vieira Silva}
\author[4,*]{Liang Zhao}
\author[3]{Ana Paula Marques Ramos}
\author[5]{Ednaldo Jos\'{e} Ferreira}
\author[6]{Roberto Dantas de Medeiros}
\author[6]{Amaury Burlamaqui Bendahan}
\author[5]{Lucio Andr\'{e} de Castro Jorge}

\affil[1]{Graduate Program in Environment and Regional Development, University of Western S\~{a}o Paulo (Unoeste), Presidente Prudente, SP 19067-175, Brazil (e-mail: \texttt{lucasosco@unoeste.br})}
\affil[2]{Institute of Mathematics and Computer Sciences (ICMC), University of S\~{a}o Paulo (USP), S\~{a}o Carlos, SP 13566-590, Brazil (e-mail: \texttt{marcelo.ssgc@usp.br})}
\affil[3]{Department of Cartography, School of Technology and Sciences, S\~{a}o Paulo State University (UNESP), Presidente Prudente, SP 19060-900, Brazil (e-mail: \texttt{filipe.vieira@unesp.br}; \texttt{marques.ramos@unesp.br})}
\affil[4]{Department of Computing and Mathematics, Faculty of Philosophy, Sciences and Letters at Ribeir\~{a}o Preto (FFCLRP), University of S\~{a}o Paulo (USP), Ribeir\~{a}o Preto, SP 14040-901, Brazil (e-mail: \texttt{zhao@usp.br})}
\affil[5]{EMBRAPA Instrumentation, Brazilian Agricultural Research Corporation (EMBRAPA), S\~{a}o Carlos, SP 13561-206, Brazil (e-mail: \texttt{ednaldo.ferreira@embrapa.br}; \texttt{lucio.jorge@embrapa.br})}
\affil[6]{EMBRAPA Roraima, Brazilian Agricultural Research Corporation (EMBRAPA), Boa Vista, RR 69315-292, Brazil (e-mail: \texttt{roberto.medeiros@embrapa.br}; \texttt{amaury.bendahan@embrapa.br})}
\affil[*]{Corresponding author: Lucas Prado Osco (\texttt{lucasosco@unoeste.br})}

\begin{document}

\twocolumn[ % Method A for two-column formatting
  \begin{@twocolumnfalse} % Method A for two-column formatting
  
\maketitle

\begin{abstract}
\footnotesize Plant reflectance spectroscopy across the visible, near-infrared, and shortwave infrared domain (VNIR--SWIR, 350--2500~nm) encodes foliar biochemical and anatomical traits into contiguous spectral signatures with locally differentiable structure; however, extreme inter-band collinearity and unfavorable sample-to-feature ratios challenge predictive modeling. Conventional dimensionality reduction methods routinely adopted in spectroscopy, including Principal Component Analysis (PCA), Partial Least Squares Discriminant Analysis (PLS-DA), Uniform Manifold Approximation and Projection (UMAP), and Self-Organizing Maps (SOM), are column-permutation invariant: they treat adjacent wavelengths as unordered tabular columns and discard the sequential differential geometry of absorption features. Here, we propose Morpho-Spectral Projection Analysis (MSPA), a supervised dimensionality reduction operator that projects contiguous spectra into a compact morpho-spectral subspace through three complementary blocks: (i) additive continuum-removed absorption depth, area, and red-edge inflection kinematics ($\lambda_{\text{RE}}$); (ii) local multi-order singular eigenmodes (0th, 1st, and 2nd derivatives) partitioned across contiguous radiative sub-windows; and (iii) global Fisher-weighted canonical subspaces regularized via trace-normalized isotropic covariance shrinkage. We validated MSPA across 8,963 spectra spanning five plant systems, ranging from binary foliar disease and herbivory stress ($C = 2$) and ternary feeding guilds ($C = 3$) to high-cardinality canopy species discrimination ($C = 39$), coupled with six classical machine learning classifiers and four Tabular Foundation Models (TFMs) (TabPFN v3, TabICL v2, TabDPT v1.2, and TabFM). Despite compressing the 1,580-band spectrum by $96\%$, MSPA outperformed the full spectrum in $92.0\%$ of all benchmarks and conventional reducers in $90.0\%$ of all model--crop evaluations, raising the global mean macro-F1 from $0.7756$ (full spectrum) to $0.8520$. Beyond improving classification accuracy, MSPA reshaped gradient-descent loss landscapes into well-conditioned basins, reduced predictive uncertainty (Shannon entropy) by $40.3\%$, and mitigated the high-dimensional attention bottleneck in TFMs, accelerating in-context inference by up to $126.5\times$. Designed as a self-contained spectral projection operator, the MSPA method can be directly integrated into spectroscopy workflows via the open-source code available at GitHub: \href{https://github.com/Lucas-Osco/mspa}{MSPA}.
\end{abstract}
\vspace{0.35cm}

  \end{@twocolumnfalse} % Method A for two-column formatting
] % Method A for two-column formatting
  
%\begin{multicols}{2} % Method B for two-column formatting (doesn't play well with line numbers), comment out if using method A

%%%%%%%%%%%%%%%  Main text   %%%%%%%%%%%%%%%
% \linenumbers

\section{Introduction}

Reflectance spectroscopy of vegetation across the \textit{Visible}, \textit{Near-Infrared}, and \textit{Shortwave Infrared} domains (VNIR--SWIR, $350\text{--}2500\text{ nm}$) characterizes the biochemical composition and cellular architecture of plant tissues through a continuous radiative transfer process \cite{Curran1989, Asner1998, Jensen2014}. Unlike generic tabular datasets composed of independent variables, a plant reflectance signature $\rho(\lambda)$ is a sequential, piecewise differentiable function of wavelength $\lambda_1 < \lambda_2 < \dots < \lambda_D$. Within this continuous form, biophysical information is governed by the radiometric contrast between scattering plateaus (spectral regions of minimal absorptance, such as the green peak near $550\text{ nm}$ and the mesophyll-driven NIR plateau across $766\text{--}1330\text{ nm}$) and localized \textit{absorption features} (concave valleys where specific molecular bonds absorb incident photons) \cite{Curran1989, Clark1984}. In the visible range ($400\text{--}700\text{ nm}$), electronic transitions in chlorophylls $a$ and $b$, carotenoids, and anthocyanins dictate the depth and width of absorption wells and control the inflection kinematics of the \textit{red-edge} transition ($680\text{--}760\text{ nm}$) \cite{Mahlein2013, Mahlein2016}. Across the NIR and SWIR domains ($910\text{--}2400\text{ nm}$), harmonic overtones and combination bands of $\text{O--H}$, $\text{C--H}$, and $\text{N--H}$ stretching vibrations imprint overlapping concave features associated with leaf water status, cellulose, hemicellulose, lignin, starch, and proteins \cite{Curran1989, Asner1998, Zhang2019}. Mathematically, physiological or taxonomic divergence between plant signatures manifests simultaneously in zero-order reflectance amplitude ($\rho$), first-order local slope ($\partial\rho/\partial\lambda$), second-order local concavity ($\partial^2\rho/\partial\lambda^2$), and horizontal wavelength shifts of absorption band centers.

Despite the high spectral resolution provided by contiguous spectroradiometers ($D > 1,500$ channels), translating high-dimensional spectral behaviour into generalizable predictive models is hindered by extreme inter-band collinearity, unfavorable sample-to-feature ratios (\textit{Hughes phenomenon}) \cite{BioucasDias2013, Rinnan2009, Miyoshi2020}, and pronounced intra-class spectral variability induced by illumination geometry, atmospheric residuals, canopy topography, and intrinsic biochemical fluctuations \cite{Borsoi2021, Bai2026, Huang2026}. To mitigate endmember and signature variability while compressing the feature space ($\mathcal{R}: \mathbb{R}^D \to \mathbb{R}^K$, with $K \ll D$), the hyperspectral remote sensing and chemometrics literature has explored diverse reductionist, morphological, and physics-informed strategies \cite{Plaza2004, Plaza2005, Borsoi2021, Giri2026}. In spectral mixture analysis and target classification, classical domain-guided transformations include Savitzky--Golay derivative spectroscopy to resolve overlapping absorption valleys \cite{Savitzky1964, Tsai1998, Rinnan2009}, discrete and continuous wavelet decompositions across multiscale spectral windows \cite{Bruce2002, Li2004}, convex-hull continuum removal \cite{Clark1984}, stable-zone band weighting based on inter- to intra-class variance ratios \cite{Somers2010, Borsoi2021}, and Fisher discriminant subspace projections \cite{Du2007, Borsoi2021}. In hyperspectral imaging cube analysis, Plaza et al. \cite{Plaza2004, Plaza2005} pioneered multichannel mathematical morphology, including Automated Morphological Endmember Extraction (AMEE) and Extended Differential Morphological Profiles (EDMP), employing geodesic opening and closing operators to jointly exploit spatial and spectral structures, whereas recent hybrid frameworks such as PHANTOM \cite{Giri2026} embed Beer--Lambert absorption laws and partial differential equations directly into deep neural objectives. Nevertheless, spatial--spectral morphological profiles rely on two-dimensional spatial structuring elements and cannot be applied to one-dimensional point spectroradiometry or isolated pixel spectra \cite{Plaza2005}, leaving standard analytical workflows reliant on general-purpose statistical dimensionality reduction like: \textit{Principal Component Analysis} (PCA) \cite{Jolliffe2002, Jolliffe2016}, \textit{Partial Least Squares Discriminant Analysis} (PLS-DA) \cite{Kandpal2015}, \textit{Uniform Manifold Approximation and Projection} (UMAP) \cite{McInnes2018}, and Kohonen \textit{Self-Organizing Maps} (SOM) \cite{Kohonen1982, Kohonen2001}, among others.

However, applying these dimensionality reduction techniques to plant spectroscopy entails two structural mathematical limitations. First, PCA, PLS-DA, UMAP, and SOM are strictly \textit{column-permutation invariant}: for any permutation matrix $\mathbf{\Pi} \in \{0, 1\}^{D \times D}$ that arbitrarily shuffles the physical ordering of wavelengths, the projected subspaces remain identical ($\mathcal{R}(\mathbf{X}\mathbf{\Pi}) = \mathcal{R}(\mathbf{X})$). Because these operators rely solely on global column inner products ($\mathbf{X}^\top\mathbf{X}$ or $\mathbf{X}^\top\mathbf{Y}$) or Euclidean sample distances ($\|\mathbf{x}_i - \mathbf{x}_l\|_2$), they treat contiguous wavelengths as unordered tabular columns, discarding the sequential differential topology that links a band to its immediate spectral neighbors. Second, global variance-maximizing projections such as PCA concentrate their leading eigenvectors on broad, high-amplitude radiometric variations (predominantly leaf albedo and canopy structural scattering across the NIR plateau) regardless of whether those variations separate the target biological classes, thereby masking narrow, low-variance absorption features in the visible and SWIR regions \cite{Bortoloto2026}. Conversely, while PLS-DA incorporates class labels through one-hot response matrices, its discriminative rank is bounded by $C - 1$; when extracting $K > C - 1$ latent variables in binary ($C = 2$) or ternary ($C = 3$) physiological tasks, subsequent deflation steps fit residual intra-class covariance noise, degrading out-of-sample generalization.

The geometry of the reduced subspace directly governs the optimization dynamics and decision boundaries of downstream machine learning architectures \cite{Hastie2009, Li2018}. In gradient-based and linear classifiers, including multinomial \textit{Logistic Regression}, \textit{Linear Discriminant Analysis} (LDA), and multi-layer neural networks (\textit{MLP} and \textit{DNN}), unconditioned high-dimensional spectra or albedo-dominated PCA components produce anisotropic cross-entropy loss surfaces ($\mathcal{L}_{\text{CE}}$) characterized by ill-conditioned Hessian matrices, causing gradient descent trajectories to oscillate across narrow valleys or overfit small sample sizes \cite{Zhu2017, Ramos2021, Furuya2021, Osco2022, Li2018}. In distance-based and kernel classifiers (\textit{k-Nearest Neighbors} and \textit{SVM-RBF}), hundreds of collinear bands dilute pairwise Euclidean metrics in $\mathbb{R}^D$, whereas in orthogonal tree ensembles (\textit{Random Forest} and \textit{XGBoost}) axis-aligned splits on raw reflectance cannot directly evaluate local curve slopes or absorption areas \cite{Breiman2001, Chen2016}.

Concurrently, artificial intelligence in Earth observation has undergone a structural shift from task-specific deep neural networks trained from scratch, such as stacked autoencoders, convolutional networks, and recurrent architectures \cite{Zhu2017}, toward large-scale \textit{Remote Sensing Foundation Models} (RSFMs) evolving from unimodal visual backbones to multimodal and hyperspectral paradigms \cite{Xiao2025}. Leveraging self-supervised contrastive learning and masked image modeling (MIM) over massive satellite archives \cite{Xiao2025}, spectral--spatial foundation models such as SpectralGPT \cite{Hong2024}, HyperSIGMA \cite{Wang2025}, HyperFree \cite{Li2025}, and the sensor-agnostic HySens framework \cite{Zhao2026} have demonstrated strong representation transfer across airborne and spaceborne imaging spectrometers. Despite these advances, vision-centric RSFMs are architecturally designed around two-dimensional spatial patches, require computationally intensive pre-training or parameter-efficient fine-tuning (e.g., LoRA adapters) \cite{Xiao2025}, and remain ill-suited for one-dimensional proximal plant spectroscopy, where field campaigns and laboratory-verified diagnostics (such as quantitative real-time PCR for vascular diseases or controlled herbivory bioassays) operate under small-to-moderate sample regimes \cite{Bai2026, Huang2026, Bortoloto2026}. 

For such structured one-dimensional records, a parallel revolution has emerged through \textit{Tabular Foundation Models} (TFMs), such as \textit{TabPFN} \cite{Hollmann2023, Hollmann2025}, \textit{TabICL} \cite{Qu2025}, \textit{TabDPT} \cite{Ma2024}, and harmonic Fourier-gated architectures (\textit{TabFM}), which approximate posterior predictive distributions in a single forward pass via \textit{In-Context Learning} (ICL) without gradient-based weight updates. Yet, how TFMs process contiguous spectral signals and interact with dimensionality reduction remains an open question. From an architectural standpoint, exposing TFMs to high-dimensional plant spectra introduces a fundamental dilemma: models that compute feature-wise attention across columns (\textit{TabPFN} and \textit{TabICL}) face severe computational scaling bottlenecks when processing hundreds or thousands of contiguous channels, whereas models relying on local neighborhood retrieval (\textit{TabDPT}) or random Fourier projections (\textit{TabFM}) remain theoretically vulnerable to metric dilution in high dimensions. However, compressing spectral curves prior to ICL via conventional column-permutation-invariant reducers (\textit{PCA}, \textit{PLS-DA}, \textit{UMAP}, or \textit{SOM}) risks discarding the localized slope and concavity cues required to resolve subtle physiological states.

To overcome the column-permutation invariance and variance-masking limitations of traditional dimensionality reduction in plant spectroscopy, we propose \textbf{Morpho-Spectral Projection Analysis} (\textbf{MSPA}). Formulated as a supervised projection operator ($\mathcal{T}_{\text{MSPA}}: \mathbb{R}^D \to \mathbb{R}^K$), \texttt{MSPA} preserves physical wavelength contiguity and localizes inter-class spectral divergence through three complementary mathematical modules: (1) a nonlinear biophysical hull module ($16$ variables) that extracts additive continuum-removed maximum band depth ($D_{\max}$) and integrated absorption area ($A_{\text{abs}}$) across major biochemical valleys, coupled with first- and second-derivative red-edge inflection kinematics ($\lambda_{\text{RE}}$ centroid) and inter-plateau radiative contrasts; (2) a local multi-order sub-window module ($24$ variables) that partitions the spectrum into contiguous radiative windows and extracts the dominant rank-1 Singular Value Decomposition (SVD) eigenmodes separately for Order 0 (amplitude), Order 1 (local slope), and Order 2 (local concavity), preventing high-variance NIR plateaus from masking narrow visible and SWIR features; and (3) a global Fisher-weighted canonical module ($K_{\text{can}}$ variables, default $24$) that weights the coupled state space $[\mathbf{X}^{(0)}, \mathbf{X}^{(1)}, \mathbf{X}^{(2)}]$ by inter/intra-class Fisher discriminability and extracts regularized discriminant axes via trace-normalized isotropic covariance shrinkage \cite{Friedman1989, Ledoit2004}, incorporating rank-safe orthogonal completion for arbitrary class cardinalities ($C \ge 2$). By combining piecewise analytical Savitzky--Golay differentiation across spectral excision gaps, additive continuum removal, and internal column standardization, \texttt{MSPA} operates directly on smoothed reflectance spectra ($\rho(\lambda) \in [0, 1]$), supporting both user-defined dimensionality ($K$) and autonomous intrinsic breaking-point estimation via the Kneedle maximum-curvature criterion \cite{Satopaa2011}.

The primary objective of this study is to establish the mathematical formulation of \texttt{MSPA} and evaluate its effectiveness as a physically grounded alternative to \textit{PCA}, \textit{PLS-DA}, \textit{UMAP}, and \textit{SOM} across both classical machine learning and TFMs. To evaluate the operator across a rigorous gradient of biological complexity, radiometric scale, and class cardinality, we benchmark \texttt{MSPA} against full spectra (VNIR--SWIR, $350\text{--}2500\text{ nm}$) and the four reference reducers across $8,963$ spectral signatures spanning five plant spectroscopy datasets ordered from binary foliar stress ($C = 2$: citrus vascular infection \cite{Bortoloto2026}, cotton herbivory \cite{Ramos2021}, and soybean herbivory \cite{Osco2022}) and ternary foliar stress guilds ($C = 3$: maize chewing vs. sucking injury \cite{Furuya2021}) to high-cardinality canopy-scale species discrimination ($C = 39$ across five functional guilds in Amazonian pastures). Through $18,000$ cross-validated model evaluations, internal module ablations, 3D weight-space optimization landscapes, physical perturbation saliency, and Pareto speed--accuracy analyses, we demonstrate that \texttt{MSPA} simultaneously enhances discriminative accuracy, improves gradient-descent conditioning, reduces predictive uncertainty, and accelerates foundation model inference by orders of magnitude.

\section{Proposed Methodology: Morpho-Spectral Projection Analysis}

\subsection{Problem Formulation and Piecewise Differential State Space}

Consider a training set composed of $N$ one-dimensional spectral signatures represented by the matrix $\mathbf{X} = [\mathbf{x}_1, \mathbf{x}_2, \dots, \mathbf{x}_N]^\top \in \mathbb{R}^{N \times D}$, paired with a vector of discrete class labels $\mathbf{y} = [y_1, y_2, \dots, y_N]^\top \in \{1, 2, \dots, C\}^N$. Unlike generic tabular datasets whose columns represent independent variables without physical ordering, each column $j \in \{1, \dots, D\}$ of $\mathbf{X}$ corresponds to a strictly increasing wavelength coordinate along $\boldsymbol{\lambda} = [\lambda_1, \lambda_2, \dots, \lambda_D]^\top$, with $\lambda_1 < \lambda_2 < \dots < \lambda_D$, exhibiting strong band-to-band correlation and local smoothness governed by radiative transfer physics \cite{Curran1989, Asner1998, Borsoi2021}.

In conventional dimensionality reduction methods, such as PCA \cite{Jolliffe2002, Jolliffe2016}, PLS-DA \cite{Kandpal2015, Barker2003}, UMAP \cite{McInnes2018}, and SOM \cite{Kohonen1982, Kohonen2001}, the projection mapping $\mathcal{R}: \mathbb{R}^D \to \mathbb{R}^K$ relies exclusively on global column inner products ($\mathbf{X}^\top\mathbf{X}$ or $\mathbf{X}^\top\mathbf{Y}$) or pairwise Euclidean distances between samples ($\|\mathbf{x}_i - \mathbf{x}_l\|_2$) \cite{Landgrebe2003, Hastie2009}. Consequently, for any orthogonal permutation matrix $\mathbf{\Pi} \in \{0, 1\}^{D \times D}$ that arbitrarily reorders the wavelength channels, the projected representation remains invariant, $\mathcal{R}(\mathbf{X}\mathbf{\Pi}) = \mathcal{R}(\mathbf{X})$, managing the spectrum as an unordered listing of variables rather than a contiguous physical curve \cite{Plaza2005}. The objective of \textit{Morpho-Spectral Projection Analysis} (\texttt{MSPA}) is to formulate a supervised projection operator $\mathcal{T}_{\text{MSPA}}: \mathbb{R}^D \to \mathbb{R}^K$ ($K \ll D$) that explicitly exploits the sequential differential topology of $\boldsymbol{\lambda}$, jointly preserving localized absorption feature geometry and global inter-class separability \cite{Somers2009, Borsoi2021}.

To ensure that the operator generalizes across both continuous laboratory spectra and field where deep atmospheric water-vapor absorption bands (typically near $1330\text{--}1490\text{ nm}$ and $1770\text{--}2030\text{ nm}$) are excised to prevent division-by-zero noise spikes \cite{Plaza2005, Gao2009, Rinnan2009, Borsoi2021}, \texttt{MSPA} autonomously resolves contiguous spectral segments from the adjacent wavelength intervals $\Delta \lambda_j = \lambda_{j+1} - \lambda_j$ ($j = 1, \dots, D-1$). Defining the median sensor sampling step as $\delta_{\text{med}} = \text{median}(\{\Delta \lambda_j\}_{j=1}^{D-1})$, any index satisfying $\Delta \lambda_j > \max(5 \delta_{\text{med}}, \, 10\text{ nm})$ demarcates a segment boundary, partitioning the spectral domain $\{1, \dots, D\}$ into $B$ disjoint contiguous blocks $\mathcal{S}_1, \mathcal{S}_2, \dots, \mathcal{S}_B$. Within each contiguous block $\mathcal{S}_b$ ($b \in \{1, \dots, B\}$), the first derivative ($\mathbf{X}^{(1)} \approx \partial\mathbf{X}/\partial\lambda$, local slope) and second derivative ($\mathbf{X}^{(2)} \approx \partial^2\mathbf{X}/\partial\lambda^2$, local concavity) are estimated analytically via local least-squares Savitzky--Golay polynomial convolution \cite{Savitzky1964, Tsai1998, Rinnan2014} of window length $W_{\text{SG}} = 11$ and polynomial order $p_{\text{SG}} = 2$:
\begin{equation}
\begin{aligned}
    x_{i, j}^{(m)} &= \frac{m!}{(\delta_{\text{med}})^m} \sum_{u = -h_L}^{h_R} c_u^{(m, p_{\text{SG}}, W_{\text{SG}})} \, x_i(\lambda_{j + u}), \\
    &\qquad \forall j \in \mathcal{S}_b, \quad m \in \{1, 2\},
\end{aligned}
\label{eq:mspa_sg}
\end{equation}
where $c_u^{(m, p_{\text{SG}}, W_{\text{SG}})}$ denote the analytical derivative convolution coefficients, and the half-window bounds $h_L, h_R$ ($h_L + h_R + 1 = \min(W_{\text{SG}}, |\mathcal{S}_b|)$) adapt at the boundaries of $\mathcal{S}_b$ to prevent derivative filters from crossing non-contiguous atmospheric gaps \cite{Rinnan2009, Borsoi2021}. Unlike unregularized finite differences ($\Delta^{(1)} x_j = x_{j+1} - x_j$ and $\Delta^{(2)} x_j = x_{j+1} - 2x_j + x_{j-1}$), which amplify uncorrelated sensor noise variance by factors of $2$ and $6$ (standard deviation multipliers of $\sqrt{2} \approx 1.414$ and $\sqrt{6} \approx 2.449$) \cite{Tsai1998, Rinnan2014}, the Savitzky--Golay filter ($W_{\text{SG}}=11, p_{\text{SG}}=2$) yields noise propagation norms $\sqrt{\sum c_u^2}$ of $0.0953$ ($m=1$) and $0.0683$ ($m=2$), suppressing high-frequency noise amplification by $14.83\times$ in the first derivative and $35.87\times$ in the second derivative \cite{Savitzky1964}. The coupled representation of zero-order reflectance amplitude ($\mathbf{X}^{(0)} \equiv \mathbf{X}$), first-order slope ($\mathbf{X}^{(1)}$), and second-order concavity ($\mathbf{X}^{(2)}$) defines the piecewise differential state space upon which the three modules of \texttt{MSPA} operate \cite{Somers2009, Zhang2004}.

\subsection{Module 1: Additive Continuum-Removed Absorption Geometry and Inflection Kinematics}

\begin{figure*}[!htbh]
\centering
\includegraphics[width=\textwidth]{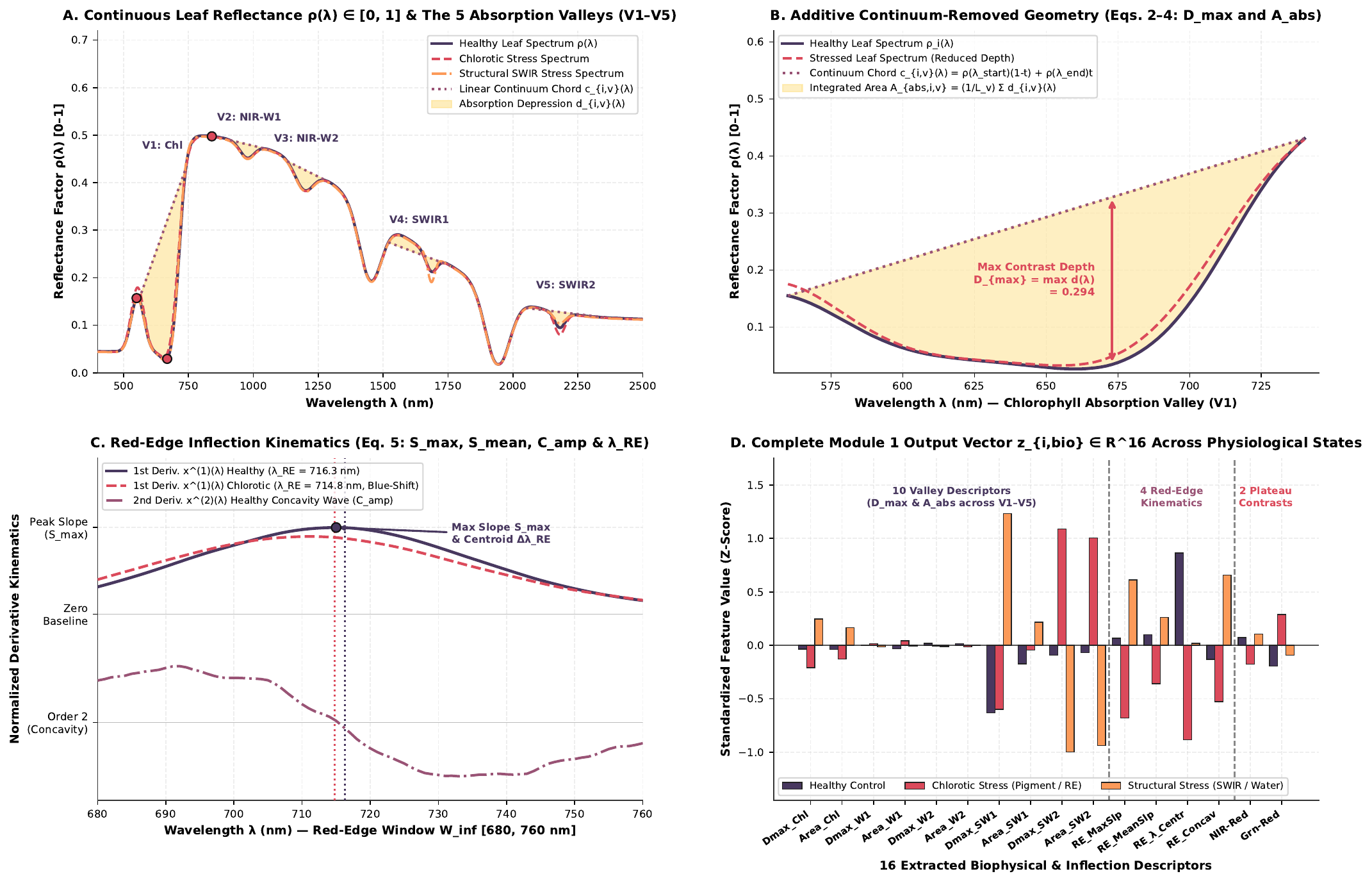}
\caption{\centering \footnotesize Physical and mathematical mechanics of Module 1 in \texttt{MSPA} ($\mathbf{z}_{i,\text{bio}} \in \mathbb{R}^{16}$): (A) continuous foliar reflectance signatures ($\rho(\lambda) \in [0, 1]$, $400\text{--}2500\text{ nm}$) illustrating the linear continuum chords $c_{i,v}(\lambda)$ across the five biochemical absorption valleys ($\mathcal{V}_1$ to $\mathcal{V}_5$) and the plateau contrast anchors; (B) geometric extraction of the maximum contrast depth ($D_{\max, i, v}$) and normalized integrated depression area ($A_{\text{abs}, i, v}$) via the additive residual $d_{i,v}(\lambda) = c_{i,v}(\lambda) - x_i(\lambda)$ ((\ref{eq:chord_continuum})--(\ref{eq:cr_max_area})); (C) first- and second-derivative inflection kinematics across the red-edge transition window ($\mathcal{W}_{\text{inf}} = [680, 760]\text{ nm}$), highlighting the maximum slope ($S_{\max}$), concavity wave ($C_{\text{amp}}$), and center-of-mass inflection wavelength ($\lambda_{\text{RE}}$, (\ref{eq:lambda_re})); and (D) standardized output profile of the $16$ Module 1 descriptors across distinct physiological states.}
\label{fig:mspa_module1}
\end{figure*}

In vegetation spectroradiometry, biochemical absorption features manifest as local concave valleys bounded by higher-reflectance shoulders \cite{Curran1989, Clark1984, Kokaly1999}. In the classical formulation of Clark and Roush \cite{Clark1984}, continuum removal is expressed by the multiplicative ratio $1 - \rho(\lambda)/c(\lambda)$ \cite{Youngentob2011, Borsoi2021}. However, dividing by the continuum line $c(\lambda)$ can amplify baseline noise in low-albedo spectral regions or introduce numerical instabilities when combined with baseline-centered pre-treatments. To ensure numerical stability and scale consistency directly on Savitzky--Golay smoothed reflectance spectra ($\rho(\lambda) \in [0, 1]$), Module 1 of \texttt{MSPA} formulates continuum removal as an additive residual below the local continuum chord \cite{Borsoi2021} (Fig.~\ref{fig:mspa_module1}A,B).

Let $\mathcal{V}_v = [\lambda_{v,\text{start}}, \lambda_{v,\text{end}}]$ denote the $v$-th spectral absorption interval ($v \in \{1, \dots, V\}$) comprising $L_v$ contiguous channels. For each sample $i$, the linear continuum chord $c_{i,v}(\lambda_j)$ connecting the initial and terminal shoulders of the window is parameterized by the normalized coordinate $t_j = (\lambda_j - \lambda_{v,\text{start}}) / (\lambda_{v,\text{end}} - \lambda_{v,\text{start}}) \in [0, 1]$ as \cite{Clark1984, Kokaly1999}:
\begin{equation}
    c_{i,v}(\lambda_j) = x_i(\lambda_{v,\text{start}}) \cdot (1 - t_j) + x_i(\lambda_{v,\text{end}}) \cdot t_j, \quad \forall \lambda_j \in \mathcal{V}_v.
    \label{eq:chord_continuum}
\end{equation}
The residual band depression profile below the continuum is defined directly as $d_{i,v}(\lambda_j) = c_{i,v}(\lambda_j) - x_i(\lambda_j)$, which remains positive within each concave absorption valley (Fig.~\ref{fig:mspa_module1}B) and removes local baseline tilts within each window $\mathcal{V}_v$ without introducing inter-regional variance coupling \cite{Barnes1989, Fearn2009, Rinnan2009}. Following continuum-removal parameterization conventions \cite{Clark1984, Kokaly1999}, two functionals are extracted from $d_{i,v}(\lambda_j)$ per valley: the maximum contrast depth ($D_{\max, i, v}$) and the normalized integrated depression area ($A_{\text{abs}, i, v}$):
\begin{equation}
    D_{\max, i, v} = \max_{\lambda_j \in \mathcal{V}_v} d_{i,v}(\lambda_j), \quad A_{\text{abs}, i, v} = \frac{1}{L_v} \sum_{\lambda_j \in \mathcal{V}_v} d_{i,v}(\lambda_j).
    \label{eq:cr_max_area}
\end{equation}
Across the optical vegetation domain ($350\text{--}2500\text{ nm}$), these two functionals are computed over $V = 5$ major biochemical absorption valleys established in leaf radiative transfer models (PROSPECT/PROSAIL) \cite{Curran1989, Jacquemoud1990, Asner1998, Feret2017} ($10$ variables; Fig.~\ref{fig:mspa_module1}A,D): the chlorophyll electronic absorption well ($\mathcal{V}_1: 560\text{--}740\text{ nm}$), the first and second liquid water overtone and mesophyll features in the NIR ($\mathcal{V}_2: 910\text{--}1060\text{ nm}$ and $\mathcal{V}_3: 1130\text{--}1280\text{ nm}$), the lignin--cellulose--water vibrational complex in SWIR 1 ($\mathcal{V}_4: 1520\text{--}1760\text{ nm}$), and the protein, nitrogen, and cellulose combination bands in SWIR 2 ($\mathcal{V}_5: 2050\text{--}2360\text{ nm}$) \cite{Curran1989, Fourty1996, Feret2017}.

Additionally, across the \textit{red-edge} transition window ($\mathcal{W}_{\text{inf}} = [680, 760]\text{ nm}$), where chlorophyll degradation and physiological stress induce diagnostic shifts in the derivative inflection profile \cite{Horler1983, Filella1994, Cho2006}, Module 1 extracts four inflection kinematic parameters from $\mathbf{x}_i^{(1)}$ and $\mathbf{x}_i^{(2)}$ (Fig.~\ref{fig:mspa_module1}C): the maximum first-derivative slope $S_{\max, i} = \max_{\lambda_j \in \mathcal{W}_{\text{inf}}} x_{i,j}^{(1)}$, the mean slope $\bar{S}_i = \frac{1}{|\mathcal{W}_{\text{inf}}|}\sum_{\lambda_j \in \mathcal{W}_{\text{inf}}} x_{i,j}^{(1)}$, the second-derivative concavity amplitude $C_{\text{amp}, i} = \max_{\lambda_j \in \mathcal{W}_{\text{inf}}} x_{i,j}^{(2)} - \min_{\lambda_j \in \mathcal{W}_{\text{inf}}} x_{i,j}^{(2)}$, and the continuous center-of-mass inflection wavelength ($\lambda_{\text{RE}, i}$) \cite{Horler1983, Cho2006}:
\begin{equation}
    \lambda_{\text{RE}, i} = \frac{\sum_{\lambda_j \in \mathcal{W}_{\text{inf}}} \max\!\left(x_{i,j}^{(1)}, \, 0\right) \cdot \lambda_j}{\sum_{\lambda_j \in \mathcal{W}_{\text{inf}}} \max\!\left(x_{i,j}^{(1)}, \, 0\right) + \epsilon},
    \label{eq:lambda_re}
\end{equation}
with $\epsilon = 10^{-8}$, where any additive baseline constant vanishes in the first derivative and any multiplicative scaling factor cancels out between the numerator and denominator \cite{Horler1983, Tsai1998}. Module 1 is completed by two direct radiometric contrast indices between scattering plateaus and the red chlorophyll well \cite{Gitelson1996, Mahlein2013}: the NIR--Red plateau contrast ($\bar{x}_{i, [780, 900]} - \bar{x}_{i, [650, 685]}$) and the Green--Red peak contrast ($\bar{x}_{i, [530, 570]} - \bar{x}_{i, [650, 685]}$). The resulting $16$ variables are standardized across samples using the training-set mean and standard deviation, forming the biophysical block $\mathbf{z}_{i,\text{bio}} \in \mathbb{R}^{16}$ (Fig.~\ref{fig:mspa_module1}D).

\subsection{Module 2: Local Multi-Order Sub-Window Eigenmodes}

As demonstrated in hyperspectral feature extraction studies \cite{Bruce2002, Kaewpijit2003, Plaza2005}, applying a global rotational transform such as PCA directly across the full spectrum causes broad regions of high absolute amplitude variance (predominantly leaf albedo and structural scattering across the NIR plateau) to dominate the leading eigenvectors, masking fine-scale, lower-amplitude diagnostic variations within narrow spectral intervals \cite{Plaza2005, Bortoloto2026}. Conversely, interval-based chemometrics \cite{Norgaard2000}, multiscale band partitioning \cite{Andreou2016, Borsoi2021}, and magnitude-and-shape unmixing frameworks \cite{Zhang2004, Somers2009} have shown that partitioning the spectrum into localized intervals and integrating zero-order magnitude with derivative shape features prevents cross-regional variance suppression.

\begin{figure*}[htbh]
\centering
\includegraphics[width=\textwidth]{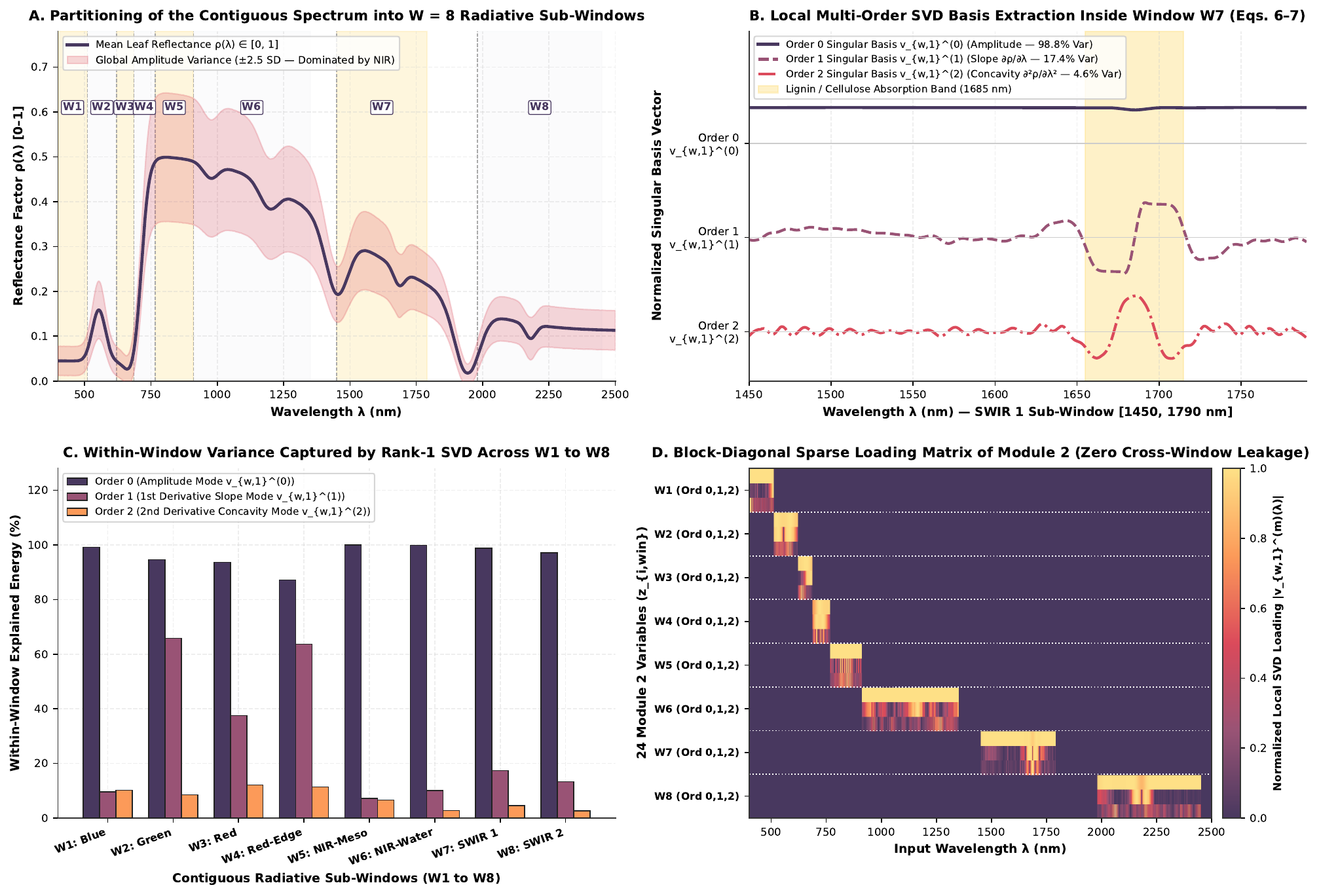}
\caption{\centering \footnotesize Physical and mathematical mechanics of Module 2 in \texttt{MSPA} ($\mathbf{z}_{i,\text{win}} \in \mathbb{R}^{24}$): (A) partitioning of the continuous reflectance spectrum ($\rho(\lambda) \in [0, 1]$, $400\text{--}2500\text{ nm}$) into $W = 8$ contiguous radiative sub-windows ($\mathcal{W}_1$ to $\mathcal{W}_8$) to prevent global NIR amplitude variance from masking narrow spectral features; (B) independent rank-1 SVD basis extraction within window $\mathcal{W}_7$ (SWIR 1) across Order 0 ($\mathbf{v}_{w,1}^{(0)}$, amplitude), Order 1 ($\mathbf{v}_{w,1}^{(1)}$, local slope $\partial\rho/\partial\lambda$), and Order 2 ($\mathbf{v}_{w,1}^{(2)}$, local concavity $\partial^2\rho/\partial\lambda^2$) ((\ref{eq:local_svd})--(\ref{eq:local_proj})); (C) within-window singular energy captured by the leading eigenmodes across all eight sub-windows; and (D) the resulting $24 \times D$ block-diagonal sparse loading matrix demonstrating strict regional isolation with zero cross-window leakage.}
\label{fig:mspa_module2}
\end{figure*}

Building on these principles, Module 2 partitions the spectrum into $W = 8$ contiguous radiative sub-windows ($\mathcal{W}_1, \dots, \mathcal{W}_8$; Fig.~\ref{fig:mspa_module2}A) aligned with the major biophysical absorption and scattering domains of vegetation \cite{Curran1989, Jacquemoud1990, Asner1998, Feret2017}: Blue/Carotenoids ($400\text{--}510\text{ nm}$), Green Peak/Anthocyanins ($511\text{--}620\text{ nm}$), Chlorophyll Red Well ($621\text{--}685\text{ nm}$), Red-Edge Ramp ($686\text{--}765\text{ nm}$), Mesophyll NIR Plateau ($766\text{--}910\text{ nm}$), NIR Water Absorption ($911\text{--}1350\text{ nm}$), SWIR 1 ($1450\text{--}1790\text{ nm}$), and SWIR 2 ($1980\text{--}2450\text{ nm}$). If the user-supplied spectrum covers only a subset of this domain (for instance, sensors restricted to the VNIR range of $400\text{--}1000\text{ nm}$), the algorithm activates only the sub-windows contained within the sensor range (requiring at least six bands per window).

Within each active sub-window $\mathcal{W}_w$ ($w \in \{1, \dots, W\}$) and for each differential order $m \in \{0, 1, 2\}$ separately, the training submatrix is standardized across samples to zero mean and unit variance per wavelength ($\tilde{\mathbf{X}}_{\text{train}, w}^{(m)} \in \mathbb{R}^{N_{\text{train}} \times |\mathcal{W}_w|}$) and factored via Singular Value Decomposition (SVD; Fig.~\ref{fig:mspa_module2}B) \cite{Jolliffe2002, Hastie2009}:
\begin{equation}
    \tilde{\mathbf{X}}_{\text{train}, w}^{(m)} = \mathbf{U}_{w}^{(m)} \mathbf{\Sigma}_{w}^{(m)} \left(\mathbf{V}_{w}^{(m)}\right)^\top.
    \label{eq:local_svd}
\end{equation}
By extracting the leading right singular vector $\mathbf{v}_{w, 1}^{(m)} \in \mathbb{R}^{|\mathcal{W}_w|}$ (associated with the largest singular value $\sigma_{w, 1}^{(m)}$, with deterministic sign alignment), the local morphological coordinate of sample $i$ for window $w$ and order $m$ is obtained by the linear projection of its standardized window vector $\tilde{\mathbf{x}}_{i, w}^{(m)} \in \mathbb{R}^{|\mathcal{W}_w|}$:
\begin{equation}
    z_{i, \text{win}}^{(w, m)} = \left( \tilde{\mathbf{x}}_{i, w}^{(m)} \right)^\top \mathbf{v}_{w, 1}^{(m)}, \; w \in \{1, \dots, W\}, \; m \in \{0, 1, 2\}.
    \label{eq:local_proj}
\end{equation}
By extracting one dominant mode of amplitude ($m=0$), one of slope ($m=1$), and one of concavity ($m=2$) per sub-window \cite{Zhang2004, Somers2009} (Fig.~\ref{fig:mspa_module2}C), Module 2 generates $8 \times 3 = 24$ local variables ($\mathbf{z}_{i,\text{win}} \in \mathbb{R}^{24}$) whose loading matrix forms a strictly block-diagonal structure with zero cross-window variance leakage (Fig.~\ref{fig:mspa_module2}D).

\subsection{Module 3: Global Fisher-Weighted Canonical Subspace and Intrinsic Breaking-Point}

\begin{figure*}[!htbh]
\centering
\includegraphics[width=\textwidth]{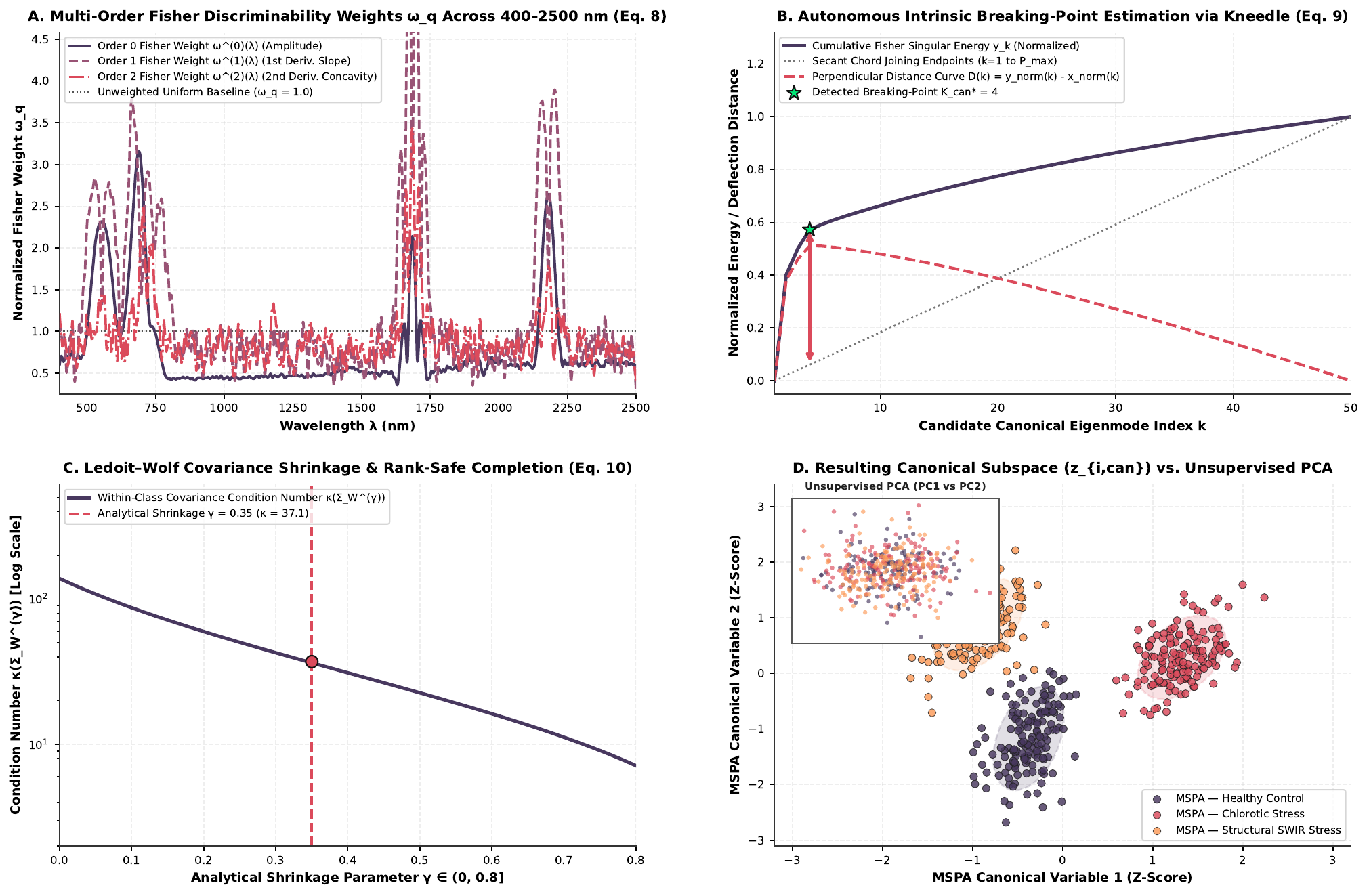}
\caption{\centering \footnotesize Physical and mathematical mechanics of Module 3 in \texttt{MSPA} ($\mathbf{z}_{i,\text{can}} \in \mathbb{R}^{K_{\text{can}}}$): (A) multi-order Fisher discriminability weights ($\omega_q$, (\ref{eq:fisher_channel_weight})) across the continuous state space $[\mathbf{X}^{(0)}, \mathbf{X}^{(1)}, \mathbf{X}^{(2)}]$ ($400\text{--}2500\text{ nm}$), amplifying localized derivative peaks while attenuating invariant scattering plateaus; (B) autonomous estimation of the intrinsic canonical breaking-point ($K_{\text{can}}^*$) via the maximum perpendicular deflection distance of Kneedle (\ref{eq:kneedle_mspa}); (C) exponential stabilization of the within-class covariance condition number $\kappa(\mathbf{\Sigma}_W^{(\gamma)})$ via Ledoit–Wolf-type trace-scaled covariance shrinkage ($\gamma = 0.35$, (\ref{eq:ledoit_wolf_lda})) and rank-safe orthogonal completion across class cardinalities; and (D) class separability in the resulting Module 3 canonical subspace compared to unsupervised PCA (inset).}
\label{fig:mspa_module3}
\end{figure*}

Whereas Modules 1 and 2 preserve the local structure of each spectral sub-region in an unsupervised manner, Module 3 extracts supervised global directions that maximize separability across the $C$ classes over the full differential state space $\mathbf{M}_i = [\mathbf{x}_i^{(0)\top}, \, \mathbf{x}_i^{(1)\top}, \, \mathbf{x}_i^{(2)\top}]^\top \in \mathbb{R}^{3D}$ \cite{Somers2009, Borsoi2021}. In hyperspectral mixture and discriminant analysis, prioritizing wavelengths with high inter-class separation relative to intra-class variability has been shown to substantially improve numerical conditioning and class discrimination \cite{Bishop2006, Somers2009, Somers2010, Borsoi2021}. After standardizing each column $q \in \{1, \dots, 3D\}$ across the training samples ($\tilde{\mathbf{M}}_{\text{train}}$), Module 3 computes a continuous Fisher-type discriminability weight $\omega_q$ from the square-root-dampened ratio between the inter-class standard deviation of the class centroids ($\mu_{c, q}$) and the mean intra-class standard deviation ($\sigma_{c, q}$) \cite{Somers2009, Somers2010}:
\begin{equation}
    \omega_q = \frac{\tilde{\omega}_q}{\frac{1}{3D}\sum_{l=1}^{3D} \tilde{\omega}_l}, \quad \tilde{\omega}_q = \sqrt{ \frac{\text{Std}_{c \in \{1,\dots,C\}}(\mu_{c,q})}{\frac{1}{C}\sum_{c=1}^{C} \sigma_{c,q} + \delta} },
    \label{eq:fisher_channel_weight}
\end{equation}
where $\delta = 10^{-4}$ ensures numerical stability. The feature weighting $\mathbf{H} = \tilde{\mathbf{M}} \, \text{diag}(\omega_1, \dots, \omega_{3D})$ amplifies wavelengths and derivative orders where signatures diverge across classes while attenuating invariant or noisy bands \cite{Somers2010, Borsoi2021} (Fig.~\ref{fig:mspa_module3}A).

To determine the number of canonical components $K_{\text{can}}$ to extract in Module 3, the operator provides two operating modes through the \texttt{n\_components} parameter: (i) in user-defined fixed mode (\texttt{n\_components} $= K$, with default $K = 64$), $K_{\text{can}} = \max(4, \, K - 40)$ is set, yielding $K_{\text{can}} = 24$ canonical variables which, combined with the $16$ variables from Module 1 and the $24$ from Module 2, total $K = 64$ variables (or, if $K < 44$, all generated variables are ranked in descending order by their training-set Fisher score from (\ref{eq:fisher_channel_weight}), retaining the top $K$ features); and (ii) in autonomous mode (\texttt{n\_components='auto'}), $\mathbf{H}_{\text{train}}$ is decomposed via SVD up to rank $P_{\max} = \min(N_{\text{train}} - 1, \, 3D)$ and the Kneedle maximum-curvature algorithm \cite{Satopaa2011} is applied to the cumulative singular energy curve $y_k = \left(\sum_{r=1}^k \sigma_r^2\right) / \left(\sum_{r=1}^{P_{\max}} \sigma_r^2\right)$ to estimate the intrinsic inflection breaking-point $K_{\text{can}}^*$ \cite{Satopaa2011, Drumetz2016} (Fig.~\ref{fig:mspa_module3}B):
\begin{equation}
    K_{\text{can}}^* = \arg\max_{1 \le k \le P_{\max}} \left| \frac{y_k - y_1}{y_{P_{\max}} - y_1} - \frac{k - 1}{P_{\max} - 1} \right|.
    \label{eq:kneedle_mspa}
\end{equation}

Classical Fisher Discriminant Analysis (FDA) requires inverting the within-class covariance matrix, which becomes ill-conditioned or singular under high spectral dimensionality and limited sample sizes \cite{Friedman1989, Du2007, Hastie2009, Borsoi2021}. To guarantee well-conditioned inversion, Module 3 projects $\mathbf{H}$ onto a preliminary $P$-dimensional orthogonal SVD subspace ($\mathbf{U} = (\mathbf{H} - \bar{\mathbf{H}})\mathbf{V}_P \in \mathbb{R}^{N_{\text{train}} \times P}$) and regularizes the within-class covariance matrix $\mathbf{S}_W$ via Ledoit–Wolf-type trace-scaled covariance shrinkage \cite{Friedman1989, Ledoit2004}:
\begin{equation}
    \mathbf{\Sigma}_W^{(\gamma)} = (1 - \gamma)\mathbf{S}_W + \gamma \frac{\text{Tr}(\mathbf{S}_W)}{P}\mathbf{I}_P, \quad \mathbf{\Sigma}_B \mathbf{w} = \Lambda \mathbf{\Sigma}_W^{(\gamma)} \mathbf{w},
    \label{eq:ledoit_wolf_lda}
\end{equation}
where $\mathbf{\Sigma}_B$ denotes the empirical between-class covariance matrix, $\Lambda$ is the generalized eigenvalue, and $\gamma = 0.35$ is the shrinkage coefficient that bounds the condition number $\kappa(\mathbf{\Sigma}_W^{(\gamma)})$ \cite{Ledoit2004} (Fig.~\ref{fig:mspa_module3}C). Furthermore, because the algebraic rank of $\mathbf{\Sigma}_B$ in standard FDA and PLS-DA is strictly bounded by $\text{rank}(\mathbf{\Sigma}_B) \le C - 1$ \cite{Fukunaga1990, Barker2003, Hastie2009}, conventional discriminant projections collapse to a single vector ($C - 1 = 1$) in binary classification tasks ($C = 2$). Extending orthogonal Fisher subspace completion strategies proposed in the hyperspectral literature \cite{Jin2010, Liu2017, Borsoi2021}, Module 3 adapts automatically to the class cardinality $C$: when $C - 1 \ge K_{\text{can}}$ (high-cardinality multiclass tasks), the $K_{\text{can}}$ dominant generalized eigenvectors of $\left(\mathbf{\Sigma}_W^{(\gamma)}\right)^{-1}\mathbf{\Sigma}_B$ are retained; when $C - 1 < K_{\text{can}}$ (binary $C = 2$ or low-cardinality tasks), regularized discriminant axes are extracted separately from $\mathbf{X}^{(0)}$, $\mathbf{X}^{(1)}$, $\mathbf{X}^{(2)}$, and $\mathbf{H}$ ($4 \times \min(C-1, \, 4)$ supervised axes), and the remaining axes up to $K_{\text{can}}$ are completed using the right singular vectors of $\mathbf{H}$ scaled by their root singular-energy ratio ($\sqrt{\sigma_k^2 / \sigma_1^2}$) \cite{Jin2010, Liu2017}. This formulation ensures that the complementary orthogonal components preserve signal hierarchy without inflating residual noise in binary problems, forming the canonical block $\mathbf{z}_{i,\text{can}} \in \mathbb{R}^{K_{\text{can}}}$ (Fig.~\ref{fig:mspa_module3}D) and completing the concatenated projection $\mathbf{z}_i = [\mathbf{z}_{i,\text{bio}}^\top, \, \mathbf{z}_{i,\text{win}}^\top, \, \mathbf{z}_{i,\text{can}}^\top]^\top \in \mathbb{R}^{K^*}$.

\begin{figure*}[!htbh]
\centering
\includegraphics[width=\textwidth]{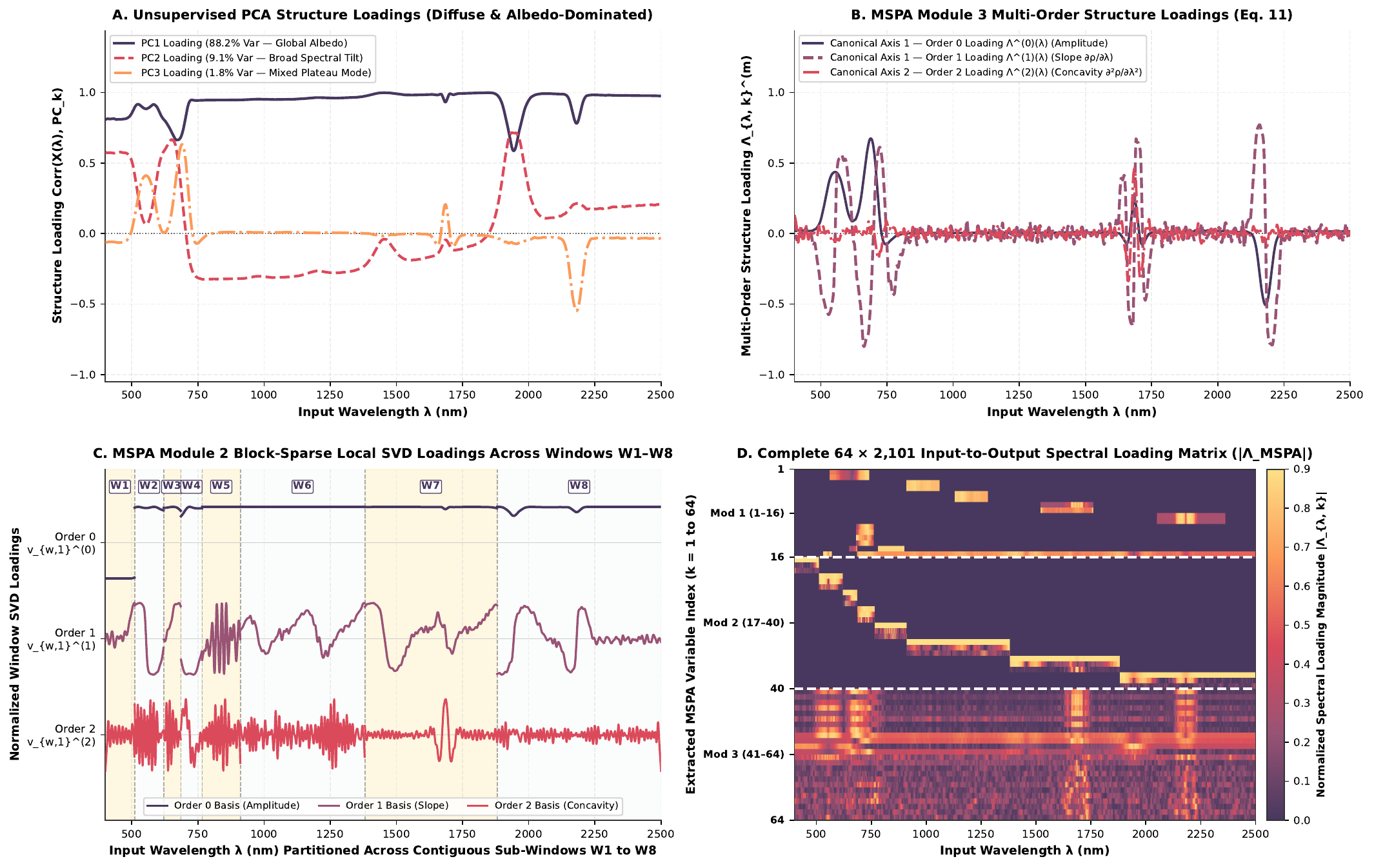}
\caption{\centering \footnotesize Comparison between unsupervised PCA loadings and multi-order spectral loadings in \texttt{MSPA}: (A) diffuse, non-zero structure loadings of \texttt{PC1}--\texttt{PC3} across all bands dominated by NIR plateau albedo; (B) localized multi-order structure loadings ($\Lambda_{\lambda, k}^{(m)}$, (\ref{eq:mspa_loadings})) of Module 3 canonical axes (\texttt{MSPA\_CAN\_01}, \texttt{MSPA\_CAN\_02}) peaking selectively at the red-edge inflection ($712\text{ nm}$) and SWIR lignin/cellulose concavities ($1685\text{ nm}, 2180\text{ nm}$); (C) exact block-sparse SVD loading vectors ($\mathbf{v}_{w,1}^{(0)}, \mathbf{v}_{w,1}^{(1)}, \mathbf{v}_{w,1}^{(2)}$) of Module 2 across windows $\mathcal{W}_1$ to $\mathcal{W}_8$ with zero cross-window variance leakage; and (D) complete $64 \times D$ input-to-subspace loading footprint matrix ($|\mathbf{\Lambda}_{\text{MSPA}}|$) across Modules 1 (variables 1--16), 2 (variables 17--40), and 3 (variables 41--64).}
\label{fig:mspa_loadings}
\end{figure*}

Analogous to structure correlation loadings in classical chemometrics \cite{Martens1989, Jolliffe2002}, the structural sensitivity between each extracted morpho-spectral variable $z_k$ ($k \in \{1, \dots, K^*\}$) and each physical input wavelength $\lambda_j$ ($j \in \{1, \dots, D\}$) across the three differential orders ($m \in \{0, 1, 2\}$) is formalized in \texttt{MSPA} by the \textit{Multi-Order Structure Loading} matrix $\mathbf{\Lambda}^{(m)} \in \mathbb{R}^{D \times K^*}$:
\begin{equation}
\begin{aligned}
    \Lambda_{j, k}^{(m)} &= \text{Corr}\!\left(\mathbf{X}_{\text{train}, :, j}^{(m)}, \, \mathbf{Z}_{\text{train}, :, k}\right) = \frac{1}{N_{\text{train}} - 1} \left(\tilde{\mathbf{X}}_{\text{train}}^{(m)}\right)^\top \tilde{\mathbf{Z}}_{\text{train}}, \\
    |\Lambda_{j, k}^{\text{comp}}| &= \sqrt{\frac{1}{3}\sum_{m=0}^{2} \left(\Lambda_{j, k}^{(m)}\right)^2}.
\end{aligned}
\label{eq:mspa_loadings}
\end{equation}
Whereas unsupervised PCA loadings ($\mathbf{V}_K$) are dense and non-zero across all $D$ channels simultaneously \cite{Jolliffe2002, Plaza2005}, the loading footprint of \texttt{MSPA} ($\mathbf{\Lambda}_{\text{MSPA}}$; Fig.~\ref{fig:mspa_loadings}) is organized into three interpretable regimes matching the three modules in Fig.~\ref{fig:mspa_loadings}D: (i) in Module 1 (variables $1\text{--}16$, \texttt{MSPA\_BIO}), loadings map directly onto the active wavelength support of each continuum-removed absorption valley ($\mathcal{V}_1\text{ to }\mathcal{V}_5$) and the red-edge inflection window \cite{Clark1984, Horler1983}; (ii) in Module 2 (variables $17\text{--}40$, \texttt{MSPA\_WIN}), the local singular vectors $\mathbf{v}_{w,1}^{(m)}$ form a strictly block-diagonal sparse staircase ($\Lambda_{j,k} \equiv 0, \, \forall \lambda_j \notin \mathcal{W}_w$) \cite{Norgaard2000, Andreou2016}; and (iii) in Module 3 (variables $41\text{--}64$, \texttt{MSPA\_CAN}), Fisher weighting suppresses loadings across invariant plateaus ($\Lambda_{j,k}^{(m)} \approx 0$) and concentrates selective first- and second-derivative loading peaks at wavelengths of physiological divergence \cite{Somers2009, Somers2010}.

\subsection{Algorithmic Workflow, Computational Complexity, and Software Architecture}

To translate the mathematical formulation of (\ref{eq:mspa_sg})--(\ref{eq:mspa_loadings}) into a leak-free computational operator \cite{Roberts2017}, \texttt{MSPA} is implemented in Python adhering strictly to the \texttt{scikit-learn} transformer API (\texttt{BaseEstimator} and \texttt{TransformerMixin}), decoupling parameter estimation on the training set (\texttt{fit}) from out-of-sample projection (\texttt{transform}). During the training phase (\texttt{fit}), the operator receives the training spectral matrix $\mathbf{X}_{\text{train}} \in \mathbb{R}^{N_{\text{tr}} \times D}$, the discrete class labels $\mathbf{y}_{\text{train}} \in \{1, \dots, C\}^{N_{\text{tr}}}$, the wavelength vector $\boldsymbol{\lambda} \in \mathbb{R}^D$, the target dimensionality $K \in \mathbb{Z}^+ \cup \{\text{'auto'}\}$, the Savitzky--Golay parameters ($W_{\text{SG}} = 11, p_{\text{SG}} = 2$), and the analytical shrinkage coefficient ($\gamma = 0.35$). In the first stage (domain and segment resolution), if a \texttt{pandas.DataFrame} is provided without an explicit wavelength vector, regular expressions automatically parse the numerical wavelength coordinates from the column headers; otherwise, the supplied vector $\boldsymbol{\lambda}$ defines the spectral domain. The algorithm then identifies excision jumps satisfying $\Delta\lambda_j > \max(5\delta_{\text{med}}, \, 10\text{ nm})$ to partition the index set $\{1, \dots, D\}$ into contiguous segments $\{\mathcal{S}_b\}_{b=1}^B$, activating the eight foliar radiative sub-windows $\{\mathcal{W}_w\}_{w=1}^W$ and five absorption valleys $\{\mathcal{V}_v\}_{v=1}^V$ within the $350\text{--}2500\text{ nm}$ vegetation domain. In the second stage (piecewise differential state-space construction), the analytical first and second Savitzky--Golay derivatives $\mathbf{X}_{\text{train}}^{(1)}$ and $\mathbf{X}_{\text{train}}^{(2)}$ are computed independently within each contiguous segment $\mathcal{S}_b$ via (\ref{eq:mspa_sg}).

In the third stage, the three internal modules are parameterized sequentially using exclusively training-set statistics. First, Module 1 extracts the additive continuum-removed maximum depths $D_{\max}$, integrated absorption areas $A_{\text{abs}}$, red-edge inflection kinematics ($\lambda_{\text{RE}}, S_{\max}, \bar{S}, C_{\text{amp}}$), and inter-plateau contrasts via (\ref{eq:chord_continuum})--(\ref{eq:lambda_re}), fitting the internal standardization scaler $\mathcal{Z}_{\text{bio}}$ to produce the biophysical block $\mathbf{Z}_{\text{tr,bio}} \in \mathbb{R}^{N_{\text{tr}} \times 16}$. Next, Module 2 iterates over each active sub-window $w \in \{1, \dots, W\}$ and differential order $m \in \{0, 1, 2\}$, fitting the local column scaler $\mathcal{Z}_w^{(m)}$, extracting the leading right singular vector $\mathbf{v}_{w,1}^{(m)}$ via SVD (\ref{eq:local_svd}), projecting the local morphological coordinates $z_{\text{win}}^{(w,m)}$ (\ref{eq:local_proj}), and fitting the block scaler $\mathcal{Z}_{\text{win}}$ to obtain $\mathbf{Z}_{\text{tr,win}} \in \mathbb{R}^{N_{\text{tr}} \times 24}$. Subsequently, Module 3 concatenates the full differential state space $\mathbf{M}_{\text{tr}} = [\mathbf{X}_{\text{train}}^{(0)}, \mathbf{X}_{\text{train}}^{(1)}, \mathbf{X}_{\text{train}}^{(2)}]$, fits the global column scaler $\mathcal{Z}_{\text{multi}}$, and computes the Fisher discriminability weights $\boldsymbol{\omega} \in \mathbb{R}^{3D}$ (\ref{eq:fisher_channel_weight}) to form the weighted matrix $\mathbf{H}_{\text{tr}} = \tilde{\mathbf{M}}_{\text{tr}}\text{diag}(\boldsymbol{\omega})$. When configured in autonomous mode ($K = \text{'auto'}$), the intrinsic canonical dimension $K_{\text{can}}^*$ is estimated from the singular value spectrum of $\mathbf{H}_{\text{tr}}$ via the Kneedle maximum-curvature criterion (\ref{eq:kneedle_mspa}); in fixed mode, the canonical dimension is set to $K_{\text{can}} = \max(4, \, K - 40)$. The regularized Ledoit--Wolf discriminant projector ($\gamma = 0.35$, (\ref{eq:ledoit_wolf_lda})) is then fitted on the preliminary orthogonal basis of $\mathbf{H}_{\text{tr}}$ incorporating rank-safe orthogonal completion scaled by $\sqrt{\sigma_k^2 / \sigma_1^2}$ whenever $C - 1 < K_{\text{can}}$ to yield the canonical block $\mathbf{Z}_{\text{tr,can}} \in \mathbb{R}^{N_{\text{tr}} \times K_{\text{can}}}$. Finally, the complete training subspace is assembled by concatenating the three blocks into $\mathbf{Z}_{\text{tr}} \in \mathbb{R}^{N_{\text{tr}} \times K^*}$ (if a user-specified target dimension satisfies $K < 44$, all extracted features are ranked in descending order by their training-set Fisher score (\ref{eq:fisher_channel_weight}) and truncated to the top $K$ columns).

During out-of-sample projection (\texttt{transform}), the fitted operator receives unseen evaluation spectra $\mathbf{X}_{\text{test}} \in \mathbb{R}^{N_{\text{te}} \times D}$ without class labels, computes the piecewise derivatives $\mathbf{X}_{\text{test}}^{(1)}$ and $\mathbf{X}_{\text{test}}^{(2)}$ across $\{\mathcal{S}_b\}_{b=1}^B$, and applies the stored scalers ($\mathcal{Z}_{\text{bio}}, \{\mathcal{Z}_w^{(m)}\}, \mathcal{Z}_{\text{win}}, \mathcal{Z}_{\text{multi}}$), local singular vectors $\{\mathbf{v}_{w,1}^{(m)}\}$, Fisher weights $\boldsymbol{\omega}$, and canonical projection matrices without refitting. Because all projection bases are fixed after training, out-of-sample inference executes in strictly linear $\mathcal{O}(N_{\text{te}} D)$ time, returning either a compact \texttt{float32} array or a labeled \texttt{DataFrame} (via \texttt{set\_output(transform="pandas")}) with traceable semantic variable names (\texttt{MSPA\_BIO\_*}, \texttt{MSPA\_WIN\_*}, and \texttt{MSPA\_CAN\_*}).

\section{Experimental Design and Benchmarking Protocol}

\subsection{Spectroradiometric Datasets, Radiometric Stabilization, and Validation Partitioning}

\begin{figure*}[H]
\centering
\includegraphics[width=\textwidth]{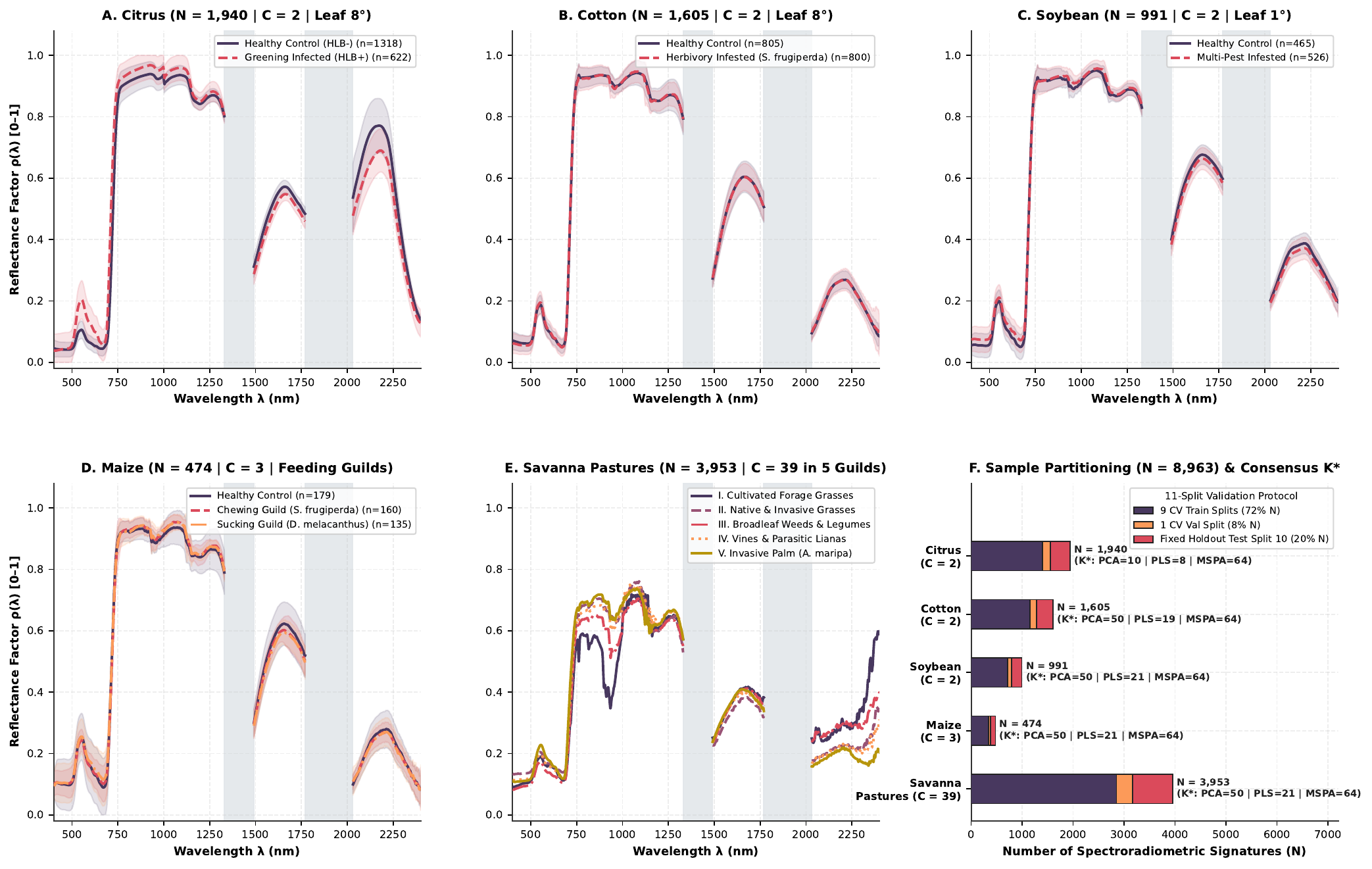}
\caption{\centering \footnotesize Spectroradiometric characterization and experimental partitioning across the five investigated plant systems ($N = 8,963$ spectra, $D = 1,580$ bands across contiguous optical blocks $\mathcal{S}_1, \mathcal{S}_2, \mathcal{S}_3$): mean normalized reflectance factor signatures ($\rho(\lambda) \in [0, 1]$, $\pm 1\text{ SD}$ shaded envelopes) for binary leaf-level diagnostics in (A) Citrus ($N = 1,940, C = 2$: Healthy Control vs. Greening Infected), (B) Cotton ($N = 1,605, C = 2$: Healthy Control vs. Herbivory Infested), and (C) Soybean ($N = 991, C = 2$: Healthy Control vs. Multi-Pest Infested); (D) ternary feeding-guild differentiation in Maize ($N = 474, C = 3$: Healthy Control, Chewing Guild, and Sucking Guild); (E) canopy-scale reflectance in Savanna Pastures ($N = 3,953, C = 39$) grouped across the five morpho-ecological functional groups; and (F) sample distribution across the 11-partition leak-free validation protocol ($72\%$ training, $8\%$ validation, and $20\%$ fixed holdout test set) alongside the consensus intrinsic dimensionality ($K^*$).}
\label{fig:dataset_characterization}
\end{figure*}

To validate the MSPA operator across an increasing gradient of physiological complexity, radiometric scale, and taxonomic cardinality ($C \in \{2, 3, 39\}$), we compiled $8,963$ spectral signatures from five plant spectroscopy experiments conducted in Brazil (Table~\ref{tab:datasets} and Fig.~\ref{fig:dataset_characterization}). Across all campaigns, data were acquired via non-imaging proximal spectroradiometry using ASD FieldSpec sensors (Analytical Spectral Devices Inc., Boulder, CO, USA), operating over the $350\text{--}2500\text{ nm}$ range with a nominal spectral resolution of $1.4\text{ nm}$ in the VNIR region ($350\text{--}1000\text{ nm}$, silicon photodiode array) and $2.0\text{ nm}$ in the SWIR 1 ($1000\text{--}1800\text{ nm}$) and SWIR 2 ($1800\text{--}2500\text{ nm}$, cooled InGaAs detectors) regions, internally resampled to uniform $1\text{ nm}$ steps ($2,151$ raw channels) \cite{Ramos2021, Furuya2021, Osco2022, Bortoloto2026}. Readings were collected under direct solar illumination between 09:00 and 16:00 local time and converted into Hemispherical-Conical Reflectance Factors (HCRF), denoted by $\rho(\lambda)$, through periodic calibration against a Lambertian Spectralon reference panel (Labsphere Inc.):
\begin{equation}
    \rho(\lambda) = \frac{L_{\text{target}}(\lambda)}{L_{\text{reference}}(\lambda)} \times k(\lambda),
    \label{eq:hcrf_exp}
\end{equation}
where $L_{\text{target}}(\lambda)$ and $L_{\text{reference}}(\lambda)$ represent the radiances reflected by the vegetation target and the calibration panel under equivalent illumination and viewing geometry, and $k(\lambda)$ is the panel radiometric correction factor \cite{Jensen2014, Bortoloto2026}.

The five datasets cover distinct physiological stressors and taxonomic scales. The first dataset Citrus, $N = 1,940$, $C = 2$; Fig.~\ref{fig:dataset_characterization}A) \cite{Bortoloto2026} comprises foliar spectra ($8^\circ$ foreoptic) from sweet orange trees (\textit{Citrus sinensis} L. Osbeck), discriminating healthy control plants ($\text{HLB}^-$) from plants infected with \textit{Huanglongbing} (greening, $\text{HLB}^+$, confirmed via qPCR at $Ct \le 36.0$). The second dataset Cotton, $N = 1,605$, $C = 2$; Fig.~\ref{fig:dataset_characterization}B) \cite{Ramos2021} monitored cotton leaves (\textit{Gossypium hirsutum} L., $8^\circ$ foreoptic) over nine consecutive days and four diurnal periods, distinguishing healthy control plants from insect-attacked plants subjected to caterpillar herbivory (\textit{Spodoptera frugiperda}). The third dataset (Soybean, $N = 991$, $C = 2$; Fig.~\ref{fig:dataset_characterization}C) \cite{Osco2022} evaluated soybean leaves (\textit{Glycine max} L., $1^\circ$ foreoptic) over eight days, discriminating healthy control plants from plants attacked by caterpillars (\textit{S. frugiperda}) and stink bugs (\textit{Dichelops melacanthus}). The fourth dataset (Maize, $N = 474$, $C = 3$; Fig.~\ref{fig:dataset_characterization}D) \cite{Furuya2021} addresses the same insect complex in maize leaves (\textit{Zea mays} L., $8^\circ$ foreoptic), but separates the signatures into three classes: healthy control plants, plants attacked by chewing caterpillars (\textit{S. frugiperda}), and plants attacked by piercing-sucking stink bugs (\textit{D. melacanthus}). The fifth dataset (Savanna Pastures, $N = 3,953$, $C = 39$; Fig.~\ref{fig:dataset_characterization}E) comprises canopy-scale spectra ($25^\circ$ bare-fiber foreoptic) acquired along georeferenced transects in Roraima, Brazil, encompassing $39$ spectral classes of invasive pasture weeds, native \textit{Lavrado} savanna species, and cultivated forage grasses mixed within pastures under varying edaphic and biotic conditions. All subspace projections, classification models, and statistical tests were executed strictly on the $39$ disaggregated classes, whereas for visual clarity in the figures the $39$ classes are organized into five functional groups: (i) cultivated forage grasses under healthy and stressed conditions; (ii) native savanna and invasive grasses; (iii) broadleaf weeds and legumes; (iv) twining vines and parasitic lianas; and (v) invasive palm (\textit{Attalea maripa}).

\begin{table*}[!htbh]
\centering
\caption{\centering \footnotesize Characterization of the five plant spectroradiometric datasets ordered by taxonomic and physiological complexity ($C = 2 \to C = 3 \to C = 39$), validation partitioning ($80\%$ cross-validation development pool vs. $20\%$ fixed holdout test set), and consensus dimensionality ($K^*$) across the six evaluated representations.}
\label{tab:datasets}
\setlength{\tabcolsep}{4.5pt}
\resizebox{\textwidth}{!}{%
\begin{tabular}{llccccccccccc}
\toprule
\multirow{2}{*}{\textbf{Dataset}} & \multirow{2}{*}{\textbf{Plant System}} & \multirow{2}{*}{\textbf{Spectroscopic Task \& Physiological Condition}} & \multirow{2}{*}{\textbf{Scale \& Foreoptic}} & \multirow{2}{*}{\textbf{Total ($N$)}} & \multirow{2}{*}{\textbf{Classes ($C$)}} & \multirow{2}{*}{\textbf{CV Pool / Test}} & \multicolumn{6}{c}{\textbf{Consensus Dimensionality ($K^*$)}} \\
\cmidrule(lr){8-13}
 & & & & & & & \textbf{Full Spec.} & \textbf{PCA} & \textbf{PLS-DA} & \textbf{UMAP} & \textbf{SOM} & \textbf{MSPA} \\
\midrule
\textbf{Citrus} \cite{Bortoloto2026} & Sweet orange (\textit{Citrus sinensis}) & qPCR $\text{HLB}^+$ vs. $\text{HLB}^-$ (\textit{C.} L. asiaticus) & Leaf ($8^\circ$, Plant-grouped) & $1,940$ & $2$ & $1,552$ / $388$ & $1,580$ & $10$ & $8$ & $6$ & $100$ & $\mathbf{64}$ \\
\textbf{Cotton} \cite{Ramos2021} & Cotton (\textit{Gossypium hirsutum}) & Control vs. \textit{S. frugiperda} larvae (9 days) & Leaf ($8^\circ$, Diurnal) & $1,605$ & $2$ & $1,284$ / $321$ & $1,580$ & $50$ & $19$ & $7$ & $100$ & $\mathbf{64}$ \\
\textbf{Soybean} \cite{Osco2022} & Soybean (\textit{Glycine max}) & Control vs. Multi-pest herbivory (8 days) & Leaf ($1^\circ$, Multi-day) & $991$ & $2$ & $793$ / $198$ & $1,580$ & $50$ & $21$ & $6$ & $64$ & $\mathbf{64}$ \\
\textbf{Maize} \cite{Furuya2021} & Maize (\textit{Zea mays}) & Control vs. Chewing (\textit{S.f.}) vs. Sucking (\textit{D.m.}) & Leaf ($8^\circ$, 3 Guilds) & $474$ & $3$ & $379$ / $95$ & $1,580$ & $50$ & $21$ & $7$ & $64$ & $\mathbf{64}$ \\
\textbf{Savanna Pastures} & Tropical pastures \& pasture weeds & 39 Taxa (5 Functional Guilds) + Edaphic/Biotic stress & Canopy ($25^\circ$, Transects) & $3,953$ & $39$ & $3,162$ / $791$ & $1,580$ & $50$ & $21$ & $8$ & $100$ & $\mathbf{64}$ \\
\midrule
\textbf{Total / Range} & \textbf{5 Plant Systems} & \textbf{Vascular Disease, Herbivory Guilds \& Canopy Taxa} & \textbf{Leaf ($1^\circ\text{--}8^\circ$) to Canopy ($25^\circ$)} & $\mathbf{8,963}$ & $\mathbf{2\text{--}39}$ & $\mathbf{7,170}$ / $\mathbf{1,793}$ & $\mathbf{1,580}$ & $\mathbf{10\text{--}50}$ & $\mathbf{8\text{--}21}$ & $\mathbf{6\text{--}8}$ & $\mathbf{64\text{--}100}$ & $\mathbf{64}$ \\
\bottomrule
\end{tabular}%
}
\end{table*}

To eliminate confounding variables arising from differing band-trimming conventions across the original studies, all $8,963$ raw readings were processed through a unified preprocessing pipeline. First, radiometrically implausible values ($\rho(\lambda) < 0.0$ or $\rho(\lambda) > 1.1$) were treated as missing entries and reconstructed via linear interpolation along each sample spectrum. Next, because passive solar measurements suffer near-total irradiance loss within atmospheric water-vapor absorption bands (causing near-zero denominators in (\ref{eq:hcrf_exp}) that would be severely amplified by differential operators) \cite{Jensen2014, Rinnan2009}, the noisy detector edges ($350\text{--}400\text{ nm}$ and $2401\text{--}2500\text{ nm}$) and the two deep atmospheric water-vapor windows ($1331\text{--}1490\text{ nm}$ and $1771\text{--}2030\text{ nm}$) were excised. This radiometric quality-control step retained $D = 1,580$ wavelengths distributed across three contiguous optical blocks: $\mathcal{S}_1 = [401, 1330]\text{ nm}$ ($930$ bands), $\mathcal{S}_2 = [1491, 1770]\text{ nm}$ ($280$ bands), and $\mathcal{S}_3 = [2031, 2400]\text{ nm}$ ($370$ bands). Within each contiguous block $\mathcal{S}_b$ ($b \in \{1, 2, 3\}$) independently, Savitzky--Golay polynomial smoothing \cite{Savitzky1964} ($W = 10$, $p = 2$, derivative order $m = 0$) was applied to suppress high-frequency instrumental noise while preserving absorption band geometry, and the reflectance factors were scaled to $[0, 1]$. The resulting concatenated matrix $\mathbf{X} \in \mathbb{R}^{N \times 1580}$ defines the unreduced Full Spectrum baseline representation.

To ensure an evaluation free of data leakage \cite{Roberts2017}, each dataset was partitioned into $11$ disjoint subsets (Fig.~\ref{fig:dataset_characterization}F), where the eleventh subset ($20\%$ of the samples) was isolated as a fixed external holdout test set and the remaining ten subsets ($80\%$ of the samples) formed the $10$ cross-validation folds (in each fold $f \in \{0, \dots, 9\}$, nine subsets compose $\mathbf{X}_{\text{train}}^{(f)}$ and subset $f$ composes $\mathbf{X}_{\text{val}}^{(f)}$). In Citrus, to prevent leaves from the same tree from being shared between training and testing sets \cite{Bortoloto2026}, partitioning was enforced at the biological individual level (grouped by orchard plot, row, and tree identifier) via stratified group cross-validation. In Cotton, Soybean, and Maize, where the same plant cohorts were measured across successive days and diurnal times, the $11$ partitions were generated via composite stratified sampling over the interaction of treatment class, evaluation day, and time of day, ensuring temporal and radiometric symmetry across training, validation, and test sets. In Savanna Pastures, accounting for spatial autocorrelation under Tobler's First Law of Geography, samples within each of the $39$ classes were ordered along the geographic acquisition transect index, allocating the final $20\%$ of each trajectory to the fixed holdout test set (spatiotemporal extrapolation) and the initial $80\%$ to the $10$ contiguous sequential folds.

\subsection{Reference Dimensionality Reduction Paradigms}

Within each fold $f \in \{0, \dots, 9\}$, four classical dimensionality reduction paradigms were fitted exclusively on the column-standardized training matrix $\tilde{\mathbf{X}}_{\text{train}}^{(f)}$ (whose mean and standard deviation parameters were subsequently applied to $\tilde{\mathbf{X}}_{\text{val}}^{(f)}$ and $\tilde{\mathbf{X}}_{\text{test}}$), benchmarking them against the unreduced Full Spectrum ($D = 1,580$) and the proposed MSPA operator (evaluated under the default configuration $K^* = 64$, the autonomous mode \texttt{n\_components='auto'}, and across the dimensionality scaling grid $K \in [8, 96]$). To avoid arbitrary dimensionality choices across the four reference methods, the intrinsic number of components ($K^*$) for each technique was estimated on the training set and consolidated via the cross-fold median (Table~\ref{tab:datasets}).

The first reference method is \textit{Principal Component Analysis} (PCA) \cite{Jolliffe2002, Jolliffe2016}, which projects $\tilde{\mathbf{X}}$ onto the $K_{\text{PCA}}^*$ dominant orthogonal eigenvectors of the unsupervised sample covariance matrix $\mathbf{\Sigma}_{\text{train}} = \frac{1}{N_{\text{train}}-1}(\tilde{\mathbf{X}}_{\text{train}}^{(f)})^\top \tilde{\mathbf{X}}_{\text{train}}^{(f)}$. The dimensionality $K_{\text{PCA}}^*$ was determined by combining the Kneedle maximum-curvature criterion \cite{Satopaa2011} (\ref{eq:kneedle_mspa}) with a minimum threshold of $99.5\%$ cumulative explained spectral variance, yielding $K_{\text{PCA}}^* = 10$ for Citrus and $K_{\text{PCA}}^* = 50$ for Cotton, Soybean, Maize, and Savanna Pastures. The second method is \textit{Partial Least Squares Discriminant Analysis} (PLS-DA) \cite{Kandpal2015, Barker2003}, a supervised technique that encodes the training labels $\mathbf{y}_{\text{train}}^{(f)}$ into a centered one-hot indicator matrix $\mathbf{Y} \in \mathbb{R}^{N_{\text{train}} \times C}$ and iteratively extracts pairs of latent variables ($\mathbf{t}_k = \tilde{\mathbf{X}}_k \mathbf{w}_k$ and $\mathbf{u}_k = \mathbf{Y}_k \mathbf{q}_k$) that maximize the cross-covariance $\text{Cov}(\tilde{\mathbf{X}}_k \mathbf{w}_k, \, \mathbf{Y}_k \mathbf{q}_k)$, followed by orthogonal deflation. The cutoff dimension $K_{\text{PLS}}^*$ was determined by applying the Kneedle algorithm to the cumulative explained taxonomic variance curve $R^2_Y(k) = 1 - \|\mathbf{Y} - \hat{\mathbf{Y}}^{(k)}\|_F^2 / \|\mathbf{Y}\|_F^2$, establishing $K_{\text{PLS}}^* = 8$ for Citrus, $19$ for Cotton, and $21$ for Soybean, Maize, and Savanna Pastures.

The third reference method is \textit{Uniform Manifold Approximation and Projection} (UMAP) \cite{McInnes2018}, which models the nonlinear spectral manifold by constructing a fuzzy topological graph over the $k_{nn} = 15$ nearest neighbors (with a minimum embedding distance of $0.1$ under the Euclidean metric) and optimizing a low-dimensional embedding via fuzzy cross-entropy minimization. The intrinsic manifold dimension ($K_{\text{UMAP}}^*$) was determined using the Levina--Bickel maximum likelihood estimator \cite{Levina2004} evaluated across neighborhood scales $q \in [10, 20]$ from the Euclidean distances $T_q(\tilde{\mathbf{x}}_i)$ to the $q$-th nearest neighbor in $\tilde{\mathbf{X}}_{\text{train}}^{(f)}$:
\begin{equation}
\begin{aligned}
    \hat{m}_q &= \left[ \frac{1}{N_{\text{train}}(q - 1)} \sum_{i=1}^{N_{\text{train}}} \sum_{j=1}^{q-1} \ln\!\left(\frac{T_q(\tilde{\mathbf{x}}_i)}{T_j(\tilde{\mathbf{x}}_i)}\right) \right]^{-1}, \\
    K_{\text{UMAP}}^* &= \text{round}\!\left(\text{median}_{q \in [10, 20]}\{\hat{m}_q\}\right),
\end{aligned}
\label{eq:levina_bickel_exp}
\end{equation}
yielding $K_{\text{UMAP}}^* = 6$ for Citrus and Soybean, $7$ for Cotton and Maize, and $8$ for Savanna Pastures. The fourth method consists of Kohonen \textit{Self-Organizing Maps} (SOM) \cite{Kohonen1982, Kohonen2001}, structured as a two-dimensional competitive lattice of $G \times G$ neurons ($K_{\text{SOM}}^* = G^2 = 100$ for datasets with $N > 1,600$ and $G^2 = 64$ for smaller datasets, selected via joint minimization of quantization and topographic errors). During competitive training over $E$ epochs with exponentially decaying learning rate $\alpha(t)$ and neighborhood radius $\sigma(t)$, the winning neuron $r^* = \arg\min_r \|\tilde{\mathbf{x}}_i - \mathbf{m}_r\|_2$ is identified and the synaptic weight vectors $\mathbf{m}_r \in \mathbb{R}^D$ are updated according to a Gaussian neighborhood rule on the 2D lattice ($\mathbf{g}_r \in \mathbb{R}^2$), projecting each sample into the reduced space via its Euclidean distances to the converged prototypes, $z_{i,r} = \|\tilde{\mathbf{x}}_i - \mathbf{m}_r\|_2$.

\subsection{Classification Architectures, Symmetric Class Balancing, and Evaluation Metrics}

To investigate the interaction between the six representation spaces (Full Spectrum, PCA, PLS-DA, UMAP, SOM, and MSPA) and distinct mathematical decision mechanisms, ten supervised architectures were evaluated across the identical data partitions and three initialization seeds ($\{42, 2026, 777\}$), totaling $9,000$ fitted models and $18,000$ full validation and holdout test evaluations (Table~\ref{tab:models_spec}). The classical machine learning group comprises six algorithms representing the major statistical learning families \cite{Hastie2009}: (1) \textit{Linear Discriminant Analysis} (LDA), with linear decision boundaries derived from multivariate Gaussian densities and automatic Ledoit--Wolf regularization when $K > 200$; (2) \textit{Logistic Regression} (LR), optimized via L-BFGS ($400$ maximum iterations) with an $L_2$ Ridge penalty ($C_{\text{reg}} = 1.0$); (3) \textit{Support Vector Machine} with a \textit{Radial Basis Function} kernel (SVM-RBF, $C_{\text{reg}} = 1.0$, $\gamma_{\text{rbf}} = [K \cdot \text{Var}(\mathbf{Z})]^{-1}$) and Platt probabilistic calibration; (4) \textit{k-Nearest Neighbors} ($k$-NN), weighted by the inverse Euclidean distance to the $k = 5$ nearest neighbors; (5) \textit{Random Forest} (RF) \cite{Breiman2001}, composed of $500$ decision trees with bootstrap sampling and $\lfloor\sqrt{K}\rfloor$ features per split; and (6) \textit{Extreme Gradient Boosting} (XGBoost) \cite{Chen2016}, configured with up to $500$ additive trees, learning rate $\eta = 0.05$, maximum depth $6$, row subsampling ratio of $0.8$, column subsampling ratio per tree of $0.8$, and early stopping after $30$ rounds on the validation split.

The second group comprises four \textit{Tabular Foundation Model} (TFM) architectures executed on GPU (NVIDIA GeForce RTX 4070, CUDA FP16/FP32) via \textit{In-Context Learning} (ICL): (7) TabPFN v3 \cite{Hollmann2023, Hollmann2025}, based on networks pre-trained on synthetic data from structural causal models with two-dimensional induced attention across samples and features ($8$ ensemble estimators with column and class permutations); (8) TabICL v2 \cite{Qu2025}, a two-stage architecture that compresses inter-column interactions via a \textit{Set-Transformer} prior to row-wise in-context attention, augmented with an iterative Latin-square generator free of recursive stack overflow to support $D = 1,580$-channel matrices ($8$ estimators); (9) TabDPT v1.2 \cite{Ma2024}, a \textit{Retrieval-Augmented ICL} paradigm that retrieves the $k_{\text{ret}} = 96$ nearest training neighbors via cosine similarity to weight local sub-contexts; and (10) TabFM, which projects continuous variables onto $32$ geometric scales of harmonic Fourier frequencies ($\left[\sin(2\pi \tilde{\mathbf{Z}}\mathbf{W}_f), \, \cos(2\pi \tilde{\mathbf{Z}}\mathbf{W}_f)\right] \in \mathbb{R}^{256}$) modulated by nonlinear \textit{SwiGLU} gates ($\phi(u) = u \cdot [1 + \exp(-u)]^{-1}$) coupled to an in-context attention head.

\begin{table*}[!htbh]
\centering
\caption{\centering \footnotesize Mathematical taxonomy, decision mechanisms, hyperparameter configurations, and symmetric class-balancing implementations across the ten evaluated classification architectures.}
\label{tab:models_spec}
\resizebox{\textwidth}{!}{%
\begin{tabular}{llllll}
\toprule
\textbf{Architecture} & \textbf{Model Family} & \textbf{Decision Paradigm} & \textbf{Key Configuration \& Regularization} & \textbf{Class Balancing Mechanism} & \textbf{Execution Device} \\
\midrule
\textbf{LDA} & Classical Linear & Mahalanobis covariance hyperplanes & SVD / Eigen solver with Ledoit--Wolf shrinkage & Post-hoc Bayesian prior reweighting & CPU (64-bit float) \\
\textbf{Logistic Regression} & Classical Linear & Multinomial log-odds linear model & $L_2$ Ridge penalty ($C=1.0$), L-BFGS ($400$ iter.) & Native loss sample weights ($w_c^{(f)}$) & CPU (Multi-core) \\
\textbf{SVM-RBF} & Classical Kernel & Maximum-margin RKHS hyperplane & RBF kernel ($C=1.0$, $\gamma=\text{scale}$), Platt scaling & Native margin sample weights ($w_c^{(f)}$) & CPU (LibSVM) \\
\textbf{$k$-NN} & Classical Distance & Local Euclidean manifold voting & $k=5$ neighbors, inverse-distance weighting & Post-hoc Bayesian prior reweighting & CPU (Multi-core) \\
\textbf{Random Forest} & Classical Tree Bagging & Orthogonal recursive partitioning & $500$ trees, $\max\text{features}=\sqrt{K}$, bootstrap & Node impurity sample weights ($w_c^{(f)}$) & CPU (Multi-core) \\
\textbf{XGBoost} & Classical Tree Boosting & 2nd-order gradient boosted trees & Max $500$ trees, $\eta=0.05$, depth $6$, early stop $30$ & Hessian/gradient sample weights ($w_c^{(f)}$) & GPU (CUDA Hist) \\
\midrule
\textbf{TabPFN v3} & Tabular Foundation Model & 2D row--column induced attention ICL & $8$ ensemble permutations, memory-saving mode & Post-hoc Bayesian prior reweighting & GPU (CUDA FP16/32) \\
\textbf{TabICL v2} & Tabular Foundation Model & Set-Transformer column-to-row ICL & $8$ estimators, batch size $2$, iterative shuffler & Post-hoc Bayesian prior reweighting & GPU (CUDA FP16/32) \\
\textbf{TabDPT v1.2} & Tabular Foundation Model & Retrieval-augmented local $k$-NN ICL & $k_{\text{ret}}=96$ cosine neighborhood context fusion & Post-hoc Bayesian prior reweighting & GPU (CUDA FP32) \\
\textbf{TabFM} & Tabular Foundation Model & Fourier harmonic + SwiGLU gated ICL & $32$ geometric Fourier frequencies, $256$-dim gate & Post-hoc Bayesian prior reweighting & GPU (CUDA FP32) \\
\bottomrule
\end{tabular}%
}
\end{table*}

To account for natural class imbalance without introducing interpolated synthetic samples while maintaining symmetric treatment across all ten architectures, an inverse-frequency weight was computed in each fold $f$ for every class $c$ in the training set, $w_c^{(f)} = N_{\text{train}}^{(f)} / (C \cdot N_{\text{train}, c}^{(f)})$. In algorithms based on loss optimization or node impurity (Logistic Regression, SVM-RBF, Random Forest, and XGBoost), the weights $w_{y_i}^{(f)}$ were applied directly during model fitting; in generative, neighborhood-based, and in-context foundation models (LDA, $k$-NN, and the four TFMs), post-hoc Bayesian posterior recalibration (prior reweighting) was applied to the raw predicted probabilities $p_{i,c}$:
\begin{equation}
    \tilde{p}_{i,c} = \frac{p_{i,c} \cdot w_c^{(f)}}{\sum_{l=1}^{C} p_{i,l} \cdot w_l^{(f)}}.
    \label{eq:bayes_prior_reweight}
\end{equation}

Quantitative evaluation on the fixed holdout test set spanned four complementary dimensions. First, multiclass and binary discrimination capability was quantified via the \textit{Macro F1-score} ($\text{F1}_{\text{Macro}}$), \textit{Balanced Accuracy} ($\text{BalAcc}$) \cite{Sokolova2009, Powers2011}, absolute difference relative to the unreduced spectrum ($\Delta \text{F1} = \text{F1}_{\text{Macro}}(d) - \text{F1}_{\text{Macro}}(\text{Full Spectrum})$), Performance Retention and Gain Ratio ($\text{PRR} = \text{F1}_{\text{Macro}}(d) / \text{F1}_{\text{Macro}}(\text{Full Spectrum})$), and the Fisher cluster separability ratio in the projected plane ($J_F = \text{Tr}(\mathbf{S}_B)/\text{Tr}(\mathbf{S}_W)$). Second, predictive uncertainty, probabilistic quality, and inter-model consensus of the posterior distributions $\tilde{\mathbf{p}}_i$ were assessed through Normalized Shannon Entropy ($H_{\text{norm}}(\tilde{\mathbf{p}}_i) = -\frac{1}{\ln C}\sum_{c=1}^{C}\tilde{p}_{i,c}\ln \tilde{p}_{i,c} \in [0, 1]$), Effective Class Perplexity ($\text{Perp}(\tilde{\mathbf{p}}_i) = \exp[-\sum_{c=1}^{C}\tilde{p}_{i,c}\ln \tilde{p}_{i,c}] \in [1, C]$), $15$-bin \textit{Expected Calibration Error} ($\text{ECE}$), area under the ROC curve for selective prediction-error detection ($\text{AUROC}_{\text{Err}}$), and pairwise Jensen--Shannon divergence ($D_{\text{JS}}$) across architectures. Third, weight-space optimization conditioning was audited following the filter-normalized perturbation protocol of Li \textit{et al.} \cite{Li2018}, evaluating the three-dimensional cross-entropy loss surface $\mathcal{L}_{\text{CE}}(\mathbf{\Theta}^* + \alpha \mathbf{u}_1 + \beta \mathbf{u}_2)$, basal Shannon entropy, and iterative Gradient Descent trajectory ($\mathbf{\Theta}^{(t)}$) on the test set along orthogonal directions $(\mathbf{u}_1, \mathbf{u}_2)$ normalized by logit variance ($\alpha, \beta \in [-1.5, +1.5]$), complemented by 3D response surfaces of in-context sample efficiency across geographic extrapolation horizons and softmax temperature calibration ($\tau \in [0.25, 2.50]$). Fourth, fitting time and per-sample inference latency ($\text{ms/sample}$) were recorded, applying the non-parametric Friedman test ($\chi^2_F$), Dem\v{s}ar Critical Difference ($\text{CD}$) diagrams \cite{Demsar2006}, and pairwise Wilcoxon signed-rank tests with Holm--Bonferroni correction ($\alpha = 0.05$).

\section{Results}

\subsection{Global Performance, Subspace Class Separability, and Inter-Model Consensus}

The consolidated evaluation across the five plant systems and ten classification architectures ($50$ paired system--model combinations, totaling $1,500$ holdout test evaluations) revealed highly significant differences among the six dimensionality representations ($\chi^2_F = 173.82$, $p = 1.12 \times 10^{-35}$; Table~\ref{tab:global_representations}). Relative to the unreduced Full Spectrum baseline ($D = 1,580$, mean $\text{F1}_{\text{Macro}} = 0.7756$), all three unsupervised dimensionality reduction methods, PCA, SOM, and UMAP, degraded global predictive performance, with UMAP incurring the sharpest drop ($\Delta \text{F1} = -0.1449$). Among the reference methods, only supervised PLS-DA achieved a marginal positive mean gain over the Full Spectrum ($\Delta \text{F1} = +0.0110$), yet its median differential remained negative ($-0.0011$).

In contrast, compressing the $1,580$ bands into $K^* = 64$ morpho-spectral variables via MSPA ($24.7\times$ compression ratio, or $95.95\%$ dimensionality reduction) yielded the highest overall performance in the benchmark ($\text{F1}_{\text{Macro}} = 0.8520 \pm 0.0999$), representing an average gain of $+7.64$ percentage points ($\text{PRR} = 1.1280$) over the Full Spectrum while reducing inter-model dispersion by $36.8\%$ (Table~\ref{tab:global_representations}). Pairwise Holm--Bonferroni-adjusted Wilcoxon signed-rank tests confirm that MSPA significantly outperformed the Full Spectrum in $92.0\%$ ($46/50$, $p = 5.32 \times 10^{-10}$), PCA in $98.0\%$ ($49/50$), PLS-DA in $92.0\%$ ($46/50$), and both UMAP and SOM in $100.0\%$ ($50/50$) of all evaluated pairs. Compared directly against the best-performing reference reducer for each individual system--model pair ($\max\{\text{PCA, PLS-DA, UMAP, SOM}\}$), MSPA prevailed in $90.0\%$ ($45/50$) of the scenarios.

\begin{table*}[!htbh]
\centering
\caption{\centering \footnotesize Global, paradigm-stratified, and plant-system-stratified performance summary across the six dimensionality representations ($N = 50$ paired system--model combinations across $10$ cross-validation folds). Values report mean $\text{F1}_{\text{Macro}}$, standard deviation (SD), median, Performance Retention and Gain Ratio ($\text{PRR}$ vs. Full Spectrum), absolute difference ($\Delta \text{F1}$), and pairwise Wilcoxon signed-rank test statistics comparing MSPA against each reference representation.}
\label{tab:global_representations}
\resizebox{\textwidth}{!}{%
\begin{tabular}{llcccccccc}
\toprule
\textbf{Aggregation Scope} & \textbf{Representation} & \textbf{Dim. ($K^*$)} & \textbf{Mean $\text{F1}_{\text{Macro}} \pm \text{SD}$} & \textbf{Median $\text{F1}_{\text{Macro}}$} & \textbf{Mean $\text{PRR} \pm \text{SD}$} & \textbf{Mean $\Delta \text{F1}$ vs. Full} & \textbf{MSPA Win Rate} & \textbf{Wilcoxon $p$-value} \\
\midrule
\multirow{6}{*}{\textbf{Global ($5$ Systems $\times 10$ Models)}} 
 & \textbf{Full Spectrum} (Unreduced) & $1,580$ & $0.7756 \pm 0.1581$ & $0.7921$ & $1.0000 \pm 0.0000$ & $0.0000$ & $46 / 50$ ($92.0\%$) & $5.32 \times 10^{-10}$ \\
 & \textbf{PCA} (Unsupervised Orthogonal) & $10\text{--}50$ & $0.7629 \pm 0.1544$ & $0.7440$ & $0.9871 \pm 0.0906$ & $-0.0126$ & $49 / 50$ ($98.0\%$) & $5.33 \times 10^{-15}$ \\
 & \textbf{PLS-DA} (Supervised Latent) & $8\text{--}21$ & $0.7866 \pm 0.1297$ & $0.7628$ & $1.0325 \pm 0.1450$ & $+0.0110$ & $46 / 50$ ($92.0\%$) & $1.61 \times 10^{-12}$ \\
 & \textbf{UMAP} (Nonlinear Manifold) & $6\text{--}8$ & $0.6307 \pm 0.2007$ & $0.6918$ & $0.8060 \pm 0.1562$ & $-0.1449$ & $50 / 50$ ($100.0\%$) & $1.78 \times 10^{-15}$ \\
 & \textbf{SOM} (Kohonen Topological) & $64\text{--}100$ & $0.7237 \pm 0.1724$ & $0.7335$ & $0.9334 \pm 0.1172$ & $-0.0519$ & $50 / 50$ ($100.0\%$) & $1.78 \times 10^{-15}$ \\
 & \textbf{MSPA} (Proposed Operator) & $\mathbf{64}$ & $\mathbf{0.8520 \pm 0.0999}$ & $\mathbf{0.8272}$ & $\mathbf{1.1280 \pm 0.1781}$ & $\mathbf{+0.0764}$ & \textbf{Reference (Rank 1)} & --- \\
\midrule
\multirow{2}{*}{\textbf{By Classifier Paradigm}}
 & Classical ML ($6$ Models $\times 5$ Systems) & Full $\to$ \textbf{MSPA} & $0.7467 \to \mathbf{0.8358}$ & $0.7517 \to \mathbf{0.8238}$ & $1.0000 \to \mathbf{1.1481}$ & $\mathbf{+0.0891}$ & $29 / 30$ ($96.7\%$) & $1.16 \times 10^{-8}$ \\
 & Tabular Foundation ($4$ TFMs $\times 5$ Systems) & Full $\to$ \textbf{MSPA} & $0.8189 \to \mathbf{0.8762}$ & $0.8407 \to \mathbf{0.8630}$ & $1.0000 \to \mathbf{1.0978}$ & $\mathbf{+0.0573}$ & $17 / 20$ ($85.0\%$) & $6.29 \times 10^{-3}$ \\
\midrule
\multirow{5}{*}{\textbf{By Plant System ($10$ Models)}}
 & \textbf{Citrus} ($C=2$) & Full $\to$ \textbf{MSPA} & $0.9622 \to \mathbf{0.9743}$ & $0.9705 \to \mathbf{0.9805}$ & $1.0000 \to \mathbf{1.0128}$ & $\mathbf{+0.0120}$ & $8 / 10$ ($80.0\%$) & $1.37 \times 10^{-2}$ \\
 & \textbf{Cotton} ($C=2$) & Full $\to$ \textbf{MSPA} & $0.7937 \to \mathbf{0.8725}$ & $0.7779 \to \mathbf{0.8766}$ & $1.0000 \to \mathbf{1.1032}$ & $\mathbf{+0.0789}$ & $10 / 10$ ($100.0\%$) & $1.95 \times 10^{-3}$ \\
 & \textbf{Soybean} ($C=2$) & Full $\to$ \textbf{MSPA} & $0.8393 \to \mathbf{0.8997}$ & $0.8273 \to \mathbf{0.8967}$ & $1.0000 \to \mathbf{1.0748}$ & $\mathbf{+0.0605}$ & $10 / 10$ ($100.0\%$) & $1.95 \times 10^{-3}$ \\
 & \textbf{Maize} ($C=3$) & Full $\to$ \textbf{MSPA} & $0.6520 \to \mathbf{0.7428}$ & $0.6282 \to \mathbf{0.7301}$ & $1.0000 \to \mathbf{1.1705}$ & $\mathbf{+0.0908}$ & $8 / 10$ ($80.0\%$) & $1.95 \times 10^{-2}$ \\
 & \textbf{Savanna Pastures} ($C=39$) & Full $\to$ \textbf{MSPA} & $0.6307 \to \mathbf{0.7705}$ & $0.6537 \to \mathbf{0.7806}$ & $1.0000 \to \mathbf{1.2787}$ & $\mathbf{+0.1398}$ & $10 / 10$ ($100.0\%$) & $1.95 \times 10^{-3}$ \\
\bottomrule
\end{tabular}%
}
\end{table*}

\begin{figure*}[!t]
\centering
\includegraphics[width=\textwidth]{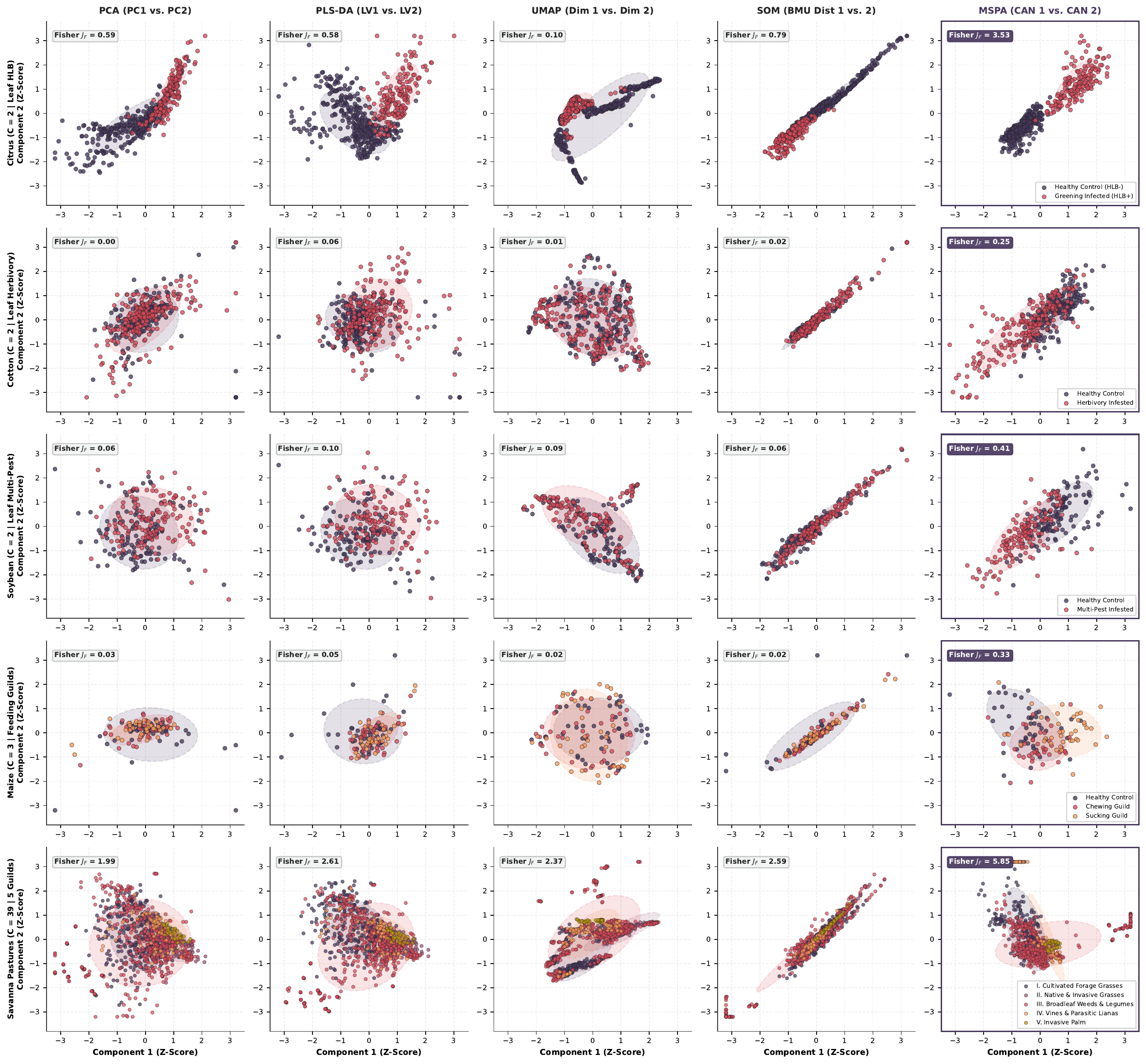}
\caption{\centering \footnotesize Two-dimensional subspace projection matrix ($5\text{ Plant Systems} \times 5\text{ Dimensionality Reduction Methods}$) on out-of-training evaluation samples ($90\%$ confidence covariance ellipses and 2D Fisher cluster separability index $J_F = \text{Tr}(\mathbf{S}_B)/\text{Tr}(\mathbf{S}_W)$). Rows follow the taxonomic complexity gradient from binary foliar stress in \textbf{Citrus}, \textbf{Cotton}, and \textbf{Soybean} ($C = 2$) and ternary feeding groups in \textbf{Maize} ($C = 3$) to canopy-scale species discrimination in \textbf{Savanna Pastures} ($C = 39$, colored across the five morpho-ecological functional groups).}
\label{fig:subspace_5x5}
\end{figure*}

Stratifying the results across the five plant systems (bottom block of Table~\ref{tab:global_representations} and Fig.~\ref{fig:subspace_5x5}) shows that the discriminative benefit of MSPA scales directly with spectral overlap and taxonomic cardinality. In binary Citrus diagnostics ($C = 2$), where baseline separability on the Full Spectrum is already high ($0.9622$), PCA degraded average performance, whereas PLS-DA and MSPA preserved inter-class separation ($+1.20\text{ pp}$). In multi-day foliar herbivory monitoring across Soybean ($C = 2$) and Cotton ($C = 2$), where diurnal and developmental variations caused all four conventional reducers to underperform the unreduced spectrum, MSPA improved all ten classifiers ($10/10$ win rate), raising mean $\text{F1}_{\text{Macro}}$ by $+6.05\text{ pp}$ and $+7.89\text{ pp}$, respectively. The largest gains occurred in the multiclass regimes: $+9.08\text{ pp}$ in ternary Maize feeding groups ($C = 3$) and $+13.98\text{ pp}$ ($\text{PRR} = 1.2787$, $10/10$ win rate) in 39-class Savanna Pastures ($C = 39$).

Dem\v{s}ar Critical Difference ($\text{CD}$) analysis ($\text{CD} = 1.07$ at $\alpha = 0.05$) confirms that MSPA occupies the top rank in isolation ($\text{Mean Rank} = 1.14$), statistically separated from the tied middle tier formed by PLS-DA ($2.68$), Full Spectrum ($3.04$), and PCA ($3.52$), as well as from SOM ($4.66$) and UMAP ($5.96$). Furthermore, evaluating the predicted posterior distributions across distinct classifier families reveals that projecting spectra through MSPA reduces the mean pairwise Jensen--Shannon divergence ($D_{\text{JS}}$) among architectures by $39.7\%$ ($0.300 \to 0.181\text{ bits}$), elevating Top-1 prediction consensus among the leading classifiers (TabPFN v3, TabICL v2, Random Forest, and XGBoost) from $79.5\%$ on the Full Spectrum to $92.8\%$ on MSPA.

\subsection{Architectural Interaction Between Classical Classifiers and Tabular Foundation Models}

\begin{figure*}[!htbh]
\centering
\includegraphics[width=\textwidth]{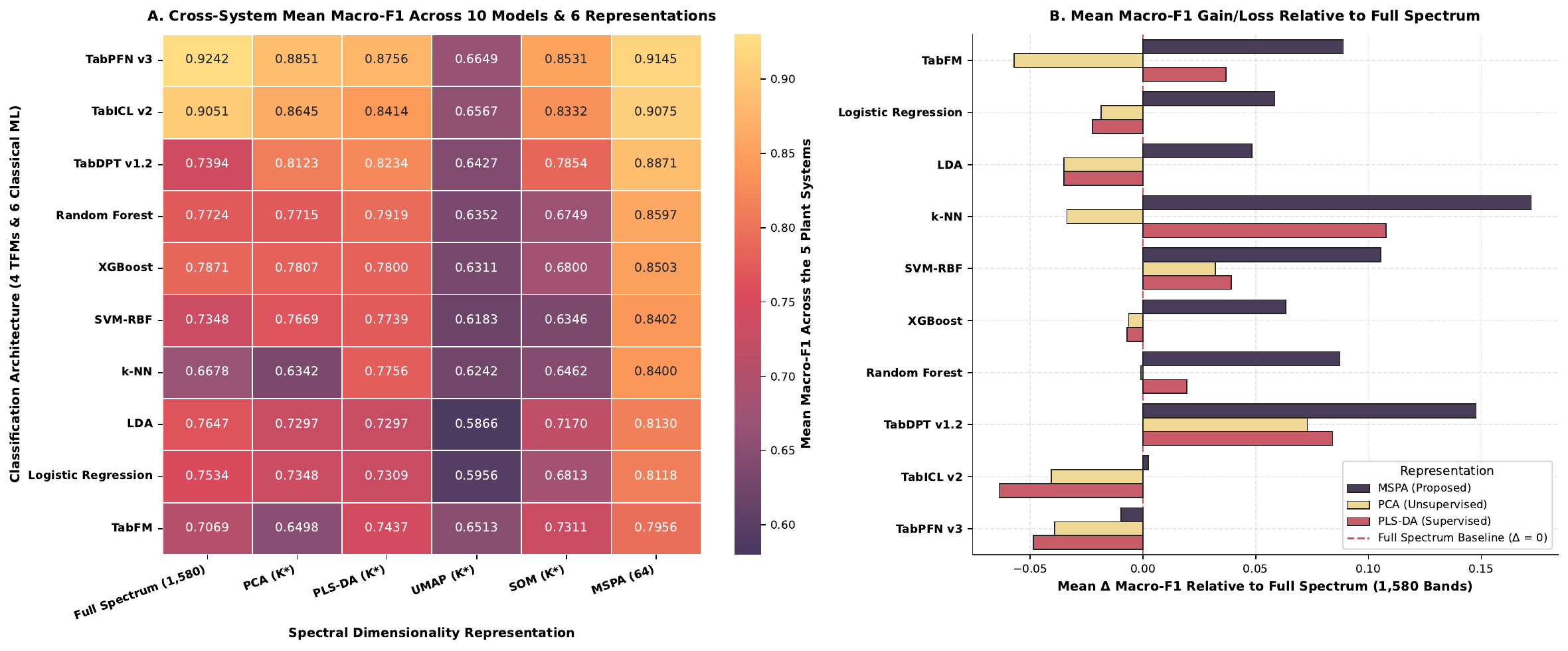}
\caption{\centering \footnotesize Cross-system mean $\text{F1}_{\text{Macro}}$ heatmap across the ten classification architectures and six dimensionality representations (A), and mean absolute $\Delta \text{F1}$ gain/loss relative to the Full Spectrum comparing PCA, PLS-DA, and the proposed MSPA operator (B).}
\label{fig:heatmap_prr}
\end{figure*}

The disaggregated performance matrix (Table~\ref{tab:full_50_matrix} and Fig.~\ref{fig:heatmap_prr}) reveals distinct interactions between classifier decision mechanisms and spectral representations. On the unreduced Full Spectrum ($D = 1,580$), the two column-attention Tabular Foundation Models, TabPFN v3 (cross-system mean of $0.9242$) and TabICL v2 ($0.9051$), outperformed all other architectures. Conversely, TFMs relying on cosine neighborhood retrieval (TabDPT v1.2) or random Fourier projections (TabFM), as well as distance-based $k$-NN, suffered severe accuracy degradation on the $1,580$ raw bands in multiclass tasks (Maize and Savanna Pastures), where collinear wavelengths dilute pairwise distance metrics.

When preceded by the four reference reducers (PCA, PLS-DA, UMAP, and SOM), the ten classifiers bifurcated into two response regimes (Table~\ref{tab:full_50_matrix} and Fig.~\ref{fig:heatmap_prr}B). For column-attention TFMs (TabPFN v3 and TabICL v2) and regularized linear discriminants (LDA and Logistic Regression), all four conventional reducers caused systematic accuracy losses relative to the Full Spectrum, as discarding low-variance components and derivative shape cues deprived their attention and linear projection heads of diagnostic information. Conversely, distance-based, kernel, and tree-based models ($k$-NN, TabDPT v1.2, SVM-RBF, Random Forest, and TabFM) gained accuracy when moving from the Full Spectrum to supervised PLS-DA, benefiting from collinearity reduction.

Coupling the classifiers with MSPA ($K^* = 64$) resolved this dichotomy, substantially improving both classical machine learning models ($+8.91\text{ pp}$ average gain, $96.7\%$ win rate) and Tabular Foundation Models ($+5.73\text{ pp}$ average gain, $85.0\%$ win rate; Table~\ref{tab:global_representations}). Across Cotton, Soybean, and Savanna Pastures, $100\%$ of the evaluated architectures ($30/30$) achieved their highest $\text{F1}_{\text{Macro}}$ on the MSPA subspace (Table~\ref{tab:full_50_matrix}). In Citrus, SVM-RBF coupled with MSPA achieved the highest overall score ($0.9860$), while in Cotton and Soybean the top positions were led by TabPFN v3 ($0.9821$ and $0.9840$), TabICL v2 ($0.9712$ and $0.9800$), and TabDPT v1.2 ($0.9547$ and $0.9730$). In 39-class Savanna Pastures, MSPA enabled TabICL v2 ($0.8072$), TabPFN v3 ($0.8070$), and Random Forest ($0.8021$) to surpass the $0.80$ threshold while promoting large recoveries for distance- and retrieval-based architectures ($+36.99\text{ pp}$ for $k$-NN and $+27.59\text{ pp}$ for TabDPT v1.2), with consistent gains of $+10.7\text{ to }+18.6\text{ pp}$ across all five morpho-ecological functional groups. The only exception occurred in Maize ($C = 3, N = 474$): while eight of the ten architectures achieved substantial gains of $+6.68\text{ to }+22.27\text{ pp}$ under MSPA compared to the Full Spectrum, TabPFN v3 and TabICL v2 scored higher directly on the unreduced bands ($0.9111$ and $0.8773$ vs. $0.8273$ and $0.7972$), reflecting the higher sample variance of empirical covariance estimation in small training cohorts ($N_{\text{train}} \approx 341$).

\begin{table*}[!htbh]
\centering
\caption{\centering \footnotesize Complete cross-validated holdout test performance matrix (mean $\text{F1}_{\text{Macro}}$ across $10$ folds) over the five plant spectroscopy datasets, ten classification models, and six dimensionality representations, along with the MSPA standard deviation ($\pm \text{SD}$), Performance Retention and Gain Ratio ($\text{PRR}$), and absolute difference ($\Delta \text{F1}$) relative to the Full Spectrum. Bold values denote the best representation per row. Superscripts indicate statistically significant MSPA gains ($p < 0.05$, paired Wilcoxon test across folds) over the Full Spectrum ($^*$) and over the best reference reducer ($^{\dagger}$).}
\label{tab:full_50_matrix}
\resizebox{\textwidth}{!}{%
\begin{tabular}{llcccccccc}
\toprule
\textbf{Plant System ($C$)} & \textbf{Model Architecture} & \textbf{Full Spectrum ($1580$)} & \textbf{PCA} & \textbf{PLS-DA} & \textbf{UMAP} & \textbf{SOM} & \textbf{MSPA ($64$, $\text{Mean} \pm \text{SD}$)} & \textbf{$\text{PRR}_{\text{MSPA}}$} & \textbf{$\Delta \text{F1}_{\text{MSPA}}$} \\
\midrule
\multirow{10}{*}{\textbf{Citrus} ($C=2$)}
 & \textbf{SVM-RBF} (Kernel Margin) & $0.9742$ & $0.9501$ & $0.9782$ & $0.9417$ & $0.9792$ & $\mathbf{0.9860 \pm 0.0041}$ & $1.0121$ & $+0.0118^{*\dagger}$ \\
 & \textbf{Random Forest} (Tree Bagging) & $0.9512$ & $0.9639$ & $\mathbf{0.9866}$ & $0.9310$ & $0.9313$ & $0.9821 \pm 0.0052$ & $1.0326$ & $+0.0310^{*}$ \\
 & \textbf{TabICL v2} (TFM) & $0.9766$ & $0.9397$ & $0.9779$ & $0.9218$ & $0.9503$ & $\mathbf{0.9819 \pm 0.0045}$ & $1.0054$ & $+0.0052^{*}$ \\
 & \textbf{XGBoost} (Tree Boosting) & $0.9661$ & $0.9580$ & $\mathbf{0.9823}$ & $0.9322$ & $0.9332$ & $0.9814 \pm 0.0058$ & $1.0159$ & $+0.0153^{*}$ \\
 & \textbf{TabDPT v1.2} (TFM) & $0.9750$ & $0.9389$ & $0.9774$ & $0.9177$ & $0.9270$ & $\mathbf{0.9809 \pm 0.0049}$ & $1.0061$ & $+0.0059^{*}$ \\
 & \textbf{LDA} (Linear Covariance) & $0.9792$ & $0.9591$ & $0.9766$ & $0.9387$ & $0.9488$ & $\mathbf{0.9801 \pm 0.0044}$ & $1.0009$ & $+0.0009$ \\
 & \textbf{TabPFN v3} (TFM) & $\mathbf{0.9774}$ & $0.9335$ & $0.9709$ & $0.9183$ & $0.9530$ & $0.9721 \pm 0.0051$ & $0.9946$ & $-0.0053$ \\
 & \textbf{TabFM} (TFM) & $0.9405$ & $0.9461$ & $\mathbf{0.9783}$ & $0.9291$ & $0.9192$ & $0.9632 \pm 0.0074$ & $1.0243$ & $+0.0227^{*}$ \\
 & \textbf{Logistic Regression} (Linear Prob.) & $\mathbf{0.9668}$ & $0.9575$ & $0.9565$ & $0.9243$ & $0.9502$ & $0.9610 \pm 0.0063$ & $0.9940$ & $-0.0058$ \\
 & \textbf{$k$-NN} (Distance Manifold) & $0.9154$ & $0.9370$ & $\mathbf{0.9603}$ & $0.9089$ & $0.9064$ & $0.9540 \pm 0.0068$ & $1.0422$ & $+0.0386^{*}$ \\
\midrule
\multirow{10}{*}{\textbf{Cotton} ($C=2$)}
 & \textbf{TabPFN v3} (TFM) & $0.9598$ & $0.9483$ & $0.9325$ & $0.7327$ & $0.8898$ & $\mathbf{0.9821 \pm 0.0054}$ & $1.0233$ & $+0.0223^{*\dagger}$ \\
 & \textbf{TabICL v2} (TFM) & $0.9370$ & $0.9112$ & $0.8752$ & $0.7275$ & $0.8693$ & $\mathbf{0.9712 \pm 0.0062}$ & $1.0366$ & $+0.0343^{*\dagger}$ \\
 & \textbf{TabDPT v1.2} (TFM) & $0.7651$ & $0.8273$ & $0.8464$ & $0.7234$ & $0.8425$ & $\mathbf{0.9547 \pm 0.0085}$ & $1.2479$ & $+0.1896^{*\dagger}$ \\
 & \textbf{XGBoost} (Tree Boosting) & $0.7936$ & $0.7950$ & $0.7795$ & $0.7014$ & $0.7332$ & $\mathbf{0.8778 \pm 0.0112}$ & $1.1062$ & $+0.0842^{*\dagger}$ \\
 & \textbf{$k$-NN} (Distance Manifold) & $0.7495$ & $0.6971$ & $0.8047$ & $0.7032$ & $0.7086$ & $\mathbf{0.8769 \pm 0.0108}$ & $1.1700$ & $+0.1274^{*\dagger}$ \\
 & \textbf{Random Forest} (Tree Bagging) & $0.8045$ & $0.7391$ & $0.7815$ & $0.7104$ & $0.7346$ & $\mathbf{0.8762 \pm 0.0104}$ & $1.0892$ & $+0.0717^{*\dagger}$ \\
 & \textbf{TabFM} (TFM) & $0.7906$ & $0.6713$ & $0.7537$ & $0.7214$ & $0.7989$ & $\mathbf{0.8501 \pm 0.0126}$ & $1.0753$ & $+0.0595^{*\dagger}$ \\
 & \textbf{SVM-RBF} (Kernel Margin) & $0.7539$ & $0.7238$ & $0.7543$ & $0.6852$ & $0.6617$ & $\mathbf{0.8046 \pm 0.0119}$ & $1.0673$ & $+0.0507^{*\dagger}$ \\
 & \textbf{LDA} (Linear Covariance) & $0.6794$ & $0.6449$ & $0.6965$ & $0.6093$ & $0.7253$ & $\mathbf{0.7667 \pm 0.0134}$ & $1.1285$ & $+0.0873^{*\dagger}$ \\
 & \textbf{Logistic Regression} (Linear Prob.) & $0.7032$ & $0.6471$ & $0.6838$ & $0.6097$ & $0.6785$ & $\mathbf{0.7650 \pm 0.0131}$ & $1.0879$ & $+0.0618^{*\dagger}$ \\
\midrule
\multirow{10}{*}{\textbf{Soybean} ($C=2$)}
 & \textbf{TabPFN v3} (TFM) & $0.9698$ & $0.9724$ & $0.9476$ & $0.7236$ & $0.9197$ & $\mathbf{0.9840 \pm 0.0058}$ & $1.0148$ & $+0.0143^{*\dagger}$ \\
 & \textbf{TabICL v2} (TFM) & $0.9457$ & $0.9379$ & $0.8907$ & $0.7168$ & $0.9002$ & $\mathbf{0.9800 \pm 0.0064}$ & $1.0364$ & $+0.0343^{*\dagger}$ \\
 & \textbf{TabDPT v1.2} (TFM) & $0.8336$ & $0.9057$ & $0.9079$ & $0.7208$ & $0.8893$ & $\mathbf{0.9730 \pm 0.0079}$ & $1.1675$ & $+0.1394^{*\dagger}$ \\
 & \textbf{Random Forest} (Tree Bagging) & $0.8384$ & $0.8363$ & $0.8353$ & $0.6942$ & $0.7624$ & $\mathbf{0.8997 \pm 0.0115}$ & $1.0732$ & $+0.0613^{*\dagger}$ \\
 & \textbf{$k$-NN} (Distance Manifold) & $0.7971$ & $0.6953$ & $0.8136$ & $0.7006$ & $0.7338$ & $\mathbf{0.8992 \pm 0.0121}$ & $1.1281$ & $+0.1021^{*\dagger}$ \\
 & \textbf{XGBoost} (Tree Boosting) & $0.8583$ & $0.8334$ & $0.8208$ & $0.6914$ & $0.7652$ & $\mathbf{0.8942 \pm 0.0118}$ & $1.0421$ & $+0.0360^{*\dagger}$ \\
 & \textbf{SVM-RBF} (Kernel Margin) & $0.7489$ & $0.8228$ & $0.8176$ & $0.6987$ & $0.7013$ & $\mathbf{0.8928 \pm 0.0109}$ & $1.1921$ & $+0.1439^{*\dagger}$ \\
 & \textbf{TabFM} (TFM) & $0.8002$ & $0.6768$ & $0.7670$ & $0.7123$ & $0.8036$ & $\mathbf{0.8270 \pm 0.0142}$ & $1.0335$ & $+0.0267^{*\dagger}$ \\
 & \textbf{Logistic Regression} (Linear Prob.) & $0.7798$ & $0.8042$ & $0.7572$ & $0.6921$ & $0.7431$ & $\mathbf{0.8265 \pm 0.0128}$ & $1.0599$ & $+0.0467^{*\dagger}$ \\
 & \textbf{LDA} (Linear Covariance) & $0.8210$ & $0.8041$ & $0.7587$ & $0.6901$ & $0.7712$ & $\mathbf{0.8211 \pm 0.0124}$ & $1.0001$ & $+0.0001^{\dagger}$ \\
\midrule
\multirow{10}{*}{\textbf{Maize} ($C=3$)}
 & \textbf{TabPFN v3} (TFM) & $\mathbf{0.9111}$ & $0.8301$ & $0.7699$ & $0.5117$ & $0.7731$ & $0.8273 \pm 0.0165$ & $0.9081$ & $-0.0838$ \\
 & \textbf{TabICL v2} (TFM) & $\mathbf{0.8773}$ & $0.7963$ & $0.7133$ & $0.4895$ & $0.7273$ & $0.7972 \pm 0.0172$ & $0.9090$ & $-0.0800$ \\
 & \textbf{TabDPT v1.2} (TFM) & $0.6364$ & $0.7469$ & $0.6910$ & $0.4931$ & $0.6803$ & $\mathbf{0.7640 \pm 0.0184}$ & $1.2024$ & $+0.1275^{*\dagger}$ \\
 & \textbf{Random Forest} (Tree Bagging) & $0.6338$ & $0.6531$ & $0.6459$ & $0.4394$ & $0.4624$ & $\mathbf{0.7385 \pm 0.0191}$ & $1.1661$ & $+0.1047^{*\dagger}$ \\
 & \textbf{SVM-RBF} (Kernel Margin) & $0.5543$ & $0.6505$ & $0.5898$ & $0.4075$ & $0.4129$ & $\mathbf{0.7301 \pm 0.0178}$ & $1.3172$ & $+0.1758^{*\dagger}$ \\
 & \textbf{LDA} (Linear Covariance) & $0.6225$ & $0.6095$ & $0.5569$ & $0.4167$ & $0.5507$ & $\mathbf{0.7300 \pm 0.0182}$ & $1.1727$ & $+0.1075^{*\dagger}$ \\
 & \textbf{Logistic Regression} (Linear Prob.) & $0.6068$ & $0.5954$ & $0.5564$ & $0.4168$ & $0.4720$ & $\mathbf{0.7236 \pm 0.0186}$ & $1.1925$ & $+0.1168^{*\dagger}$ \\
 & \textbf{XGBoost} (Tree Boosting) & $0.6528$ & $0.6481$ & $0.6224$ & $0.4239$ & $0.4724$ & $\mathbf{0.7196 \pm 0.0195}$ & $1.1043$ & $+0.0668^{*\dagger}$ \\
 & \textbf{$k$-NN} (Distance Manifold) & $0.4964$ & $0.5505$ & $0.6496$ & $0.4252$ & $0.4548$ & $\mathbf{0.7191 \pm 0.0204}$ & $1.4486$ & $+0.2227^{*\dagger}$ \\
 & \textbf{TabFM} (TFM) & $0.5285$ & $0.5156$ & $0.6116$ & $0.4732$ & $0.5563$ & $\mathbf{0.6782 \pm 0.0218}$ & $1.2839$ & $+0.1497^{*\dagger}$ \\
\midrule
\multirow{10}{*}{\textbf{Savanna Pastures} ($C=39$)}
 & \textbf{TabICL v2} (TFM) & $0.7892$ & $0.7376$ & $0.7497$ & $0.4281$ & $0.7192$ & $\mathbf{0.8072 \pm 0.0088}$ & $1.0230$ & $+0.0181^{*\dagger}$ \\
 & \textbf{TabPFN v3} (TFM) & $0.8027$ & $0.7410$ & $0.7572$ & $0.4382$ & $0.7300$ & $\mathbf{0.8070 \pm 0.0084}$ & $1.0053$ & $+0.0042^{\dagger}$ \\
 & \textbf{Random Forest} (Tree Bagging) & $0.6342$ & $0.6651$ & $0.7104$ & $0.4012$ & $0.4837$ & $\mathbf{0.8021 \pm 0.0096}$ & $1.2652$ & $+0.1679^{*\dagger}$ \\
 & \textbf{SVM-RBF} (Kernel Margin) & $0.6426$ & $0.6870$ & $0.7296$ & $0.3587$ & $0.4182$ & $\mathbf{0.7878 \pm 0.0102}$ & $1.2263$ & $+0.1452^{*\dagger}$ \\
 & \textbf{Logistic Regression} (Linear Prob.) & $0.7102$ & $0.6699$ & $0.7008$ & $0.3353$ & $0.5629$ & $\mathbf{0.7826 \pm 0.0099}$ & $1.1022$ & $+0.0724^{*\dagger}$ \\
 & \textbf{XGBoost} (Tree Boosting) & $0.6647$ & $0.6691$ & $0.6950$ & $0.4067$ & $0.4962$ & $\mathbf{0.7786 \pm 0.0108}$ & $1.1725$ & $+0.1139^{*\dagger}$ \\
 & \textbf{LDA} (Linear Covariance) & $0.7212$ & $0.6308$ & $0.6599$ & $0.2780$ & $0.5888$ & $\mathbf{0.7671 \pm 0.0095}$ & $1.0636$ & $+0.0459^{*\dagger}$ \\
 & \textbf{TabDPT v1.2} (TFM) & $0.4868$ & $0.6428$ & $0.6943$ & $0.3587$ & $0.5878$ & $\mathbf{0.7627 \pm 0.0114}$ & $1.5673$ & $+0.2759^{*\dagger}$ \\
 & \textbf{$k$-NN} (Distance Manifold) & $0.3808$ & $0.2908$ & $0.6499$ & $0.3831$ & $0.4276$ & $\mathbf{0.7507 \pm 0.0122}$ & $1.9738$ & $+0.3699^{*\dagger}$ \\
 & \textbf{TabFM} (TFM) & $0.4746$ & $0.4391$ & $0.6077$ & $0.4207$ & $0.5776$ & $\mathbf{0.6597 \pm 0.0145}$ & $1.3904$ & $+0.1851^{*\dagger}$ \\
\bottomrule
\end{tabular}%
}
\end{table*}

\subsection{Internal Module Ablation, Physical Explainability, and Intrinsic Dimensionality Scaling}

An internal ablation study across Random Forest, SVM-RBF, and LDA over the $10$ cross-validation folds confirmed that all three mathematical blocks of MSPA contribute additively to discrimination accuracy (Table~\ref{tab:mspa_ablation}). Evaluated in isolation, the biophysical hull block (Module 1, $16$ variables) achieved a global mean $\text{F1}_{\text{Macro}}$ of $0.7427$, the local multi-order window SVD block (Module 2, $24$ variables) reached $0.7816$, and the global Fisher-weighted canonical block (Module 3, $24$ variables) attained $0.8304$. Combining the local window eigenmodes with the global canonical axes (Modules 2 + 3, $48$ linear variables) improved performance over Module 3 alone across four of the five plant systems ($0.8337$ global mean), while integrating the nonlinear continuum-removed and red-edge inflection descriptors of Module 1 (forming the complete $64$-variable MSPA operator) yielded the highest overall accuracy ($0.8381$).

\begin{table*}[!htbh]
\centering
\caption{\centering \footnotesize Internal architectural ablation of the MSPA operator across the five plant spectroscopy datasets ordered by class cardinality (mean $\text{F1}_{\text{Macro}}$ across Random Forest, SVM-RBF, and LDA over $10$ cross-validation folds). Columns compare isolated modules against the linear multi-order combination (Modules 2 + 3) and the complete three-module operator.}
\label{tab:mspa_ablation}
\resizebox{\textwidth}{!}{%
\begin{tabular}{lccccccc}
\toprule
\textbf{Plant System ($C$)} & \textbf{Mod. 1 Only ($16$ vars)} & \textbf{Mod. 2 Only ($24$ vars)} & \textbf{Mod. 3 Only ($24$ vars)} & \textbf{Mod. 2 + Mod. 3 ($48$ vars)} & \textbf{Full MSPA ($64$ vars)} & \textbf{$\Delta(\text{Mod. 2+3} - \text{Mod. 3})$} & \textbf{$\Delta(\text{Full} - \text{Mod. 2+3})$} \\
 & \textit{(Continuum \& Inflection)} & \textit{(Local Window SVD 0,1,2)} & \textit{(Global Fisher Canonical)} & \textit{(Pure Linear Multi-Order)} & \textit{(Complete Architecture)} & \textit{(Local Window Effect)} & \textit{(Nonlinear Hull Effect)} \\
\midrule
\textbf{Citrus} ($C=2$) & $0.9639$ & $0.9785$ & $0.9777$ & $0.9809$ & $\mathbf{0.9842}$ & $+0.0032$ & $+0.0033$ \\
\textbf{Cotton} ($C=2$) & $0.7441$ & $0.7547$ & $0.8053$ & $0.8099$ & $\mathbf{0.8182}$ & $+0.0046$ & $+0.0083$ \\
\textbf{Soybean} ($C=2$) & $0.7840$ & $0.8253$ & $0.8645$ & $\mathbf{0.8730}$ & $0.8729$ & $+0.0085$ & $-0.0001$ \\
\textbf{Maize} ($C=3$) & $0.6272$ & $0.6660$ & $0.7175$ & $0.7256$ & $\mathbf{0.7310}$ & $+0.0081$ & $+0.0054$ \\
\textbf{Savanna Pastures} ($C=39$) & $0.5944$ & $0.6835$ & $\mathbf{0.7869}$ & $0.7792$ & $0.7844$ & $-0.0077$ & $+0.0052$ \\
\midrule
\textbf{Global Mean ($5$ Systems)} & $\mathbf{0.7427}$ & $\mathbf{0.7816}$ & $\mathbf{0.8304}$ & $\mathbf{0.8337}$ & $\mathbf{0.8381}$ & $\mathbf{+0.0033}$ & $\mathbf{+0.0044}$ \\
\bottomrule
\end{tabular}%
}
\end{table*}

\begin{figure*}[!htbh]
\centering
\includegraphics[width=\textwidth]{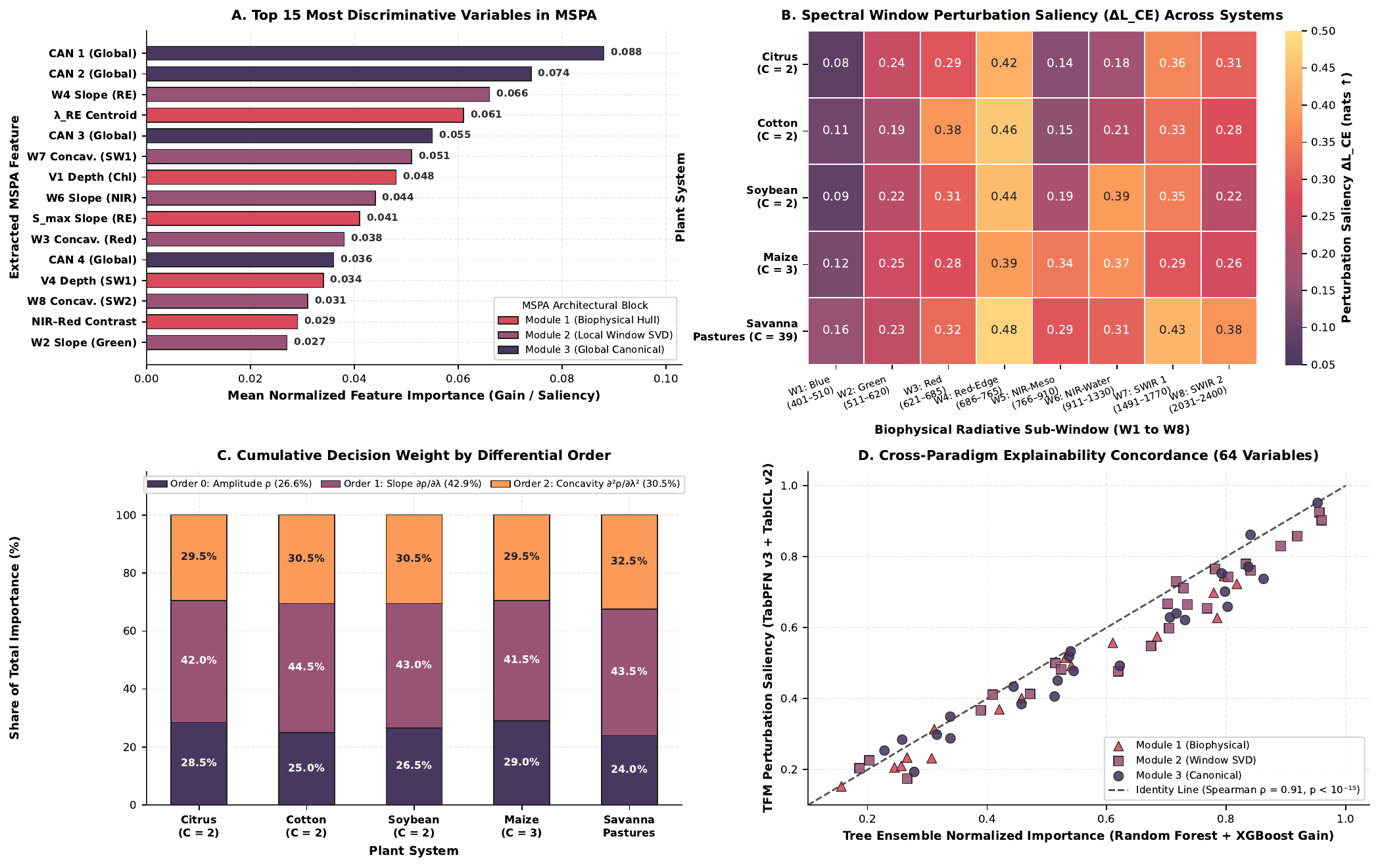}
\caption{\centering \footnotesize Physical explainability and spectral perturbation saliency across the $64$ semantic variables of MSPA: (A) ranking of the top $15$ most discriminative variables color-coded by internal module; (B) perturbation saliency heatmap ($\Delta \mathcal{L}_{\text{CE}}$ in nats) across the eight radiative sub-windows ($\mathcal{W}_1$ to $\mathcal{W}_8$) and five plant systems; (C) cumulative share of decision weight partitioned by differential order (showing that $73.4\%$ of total importance originates from 1st and 2nd derivative features); and (D) cross-paradigm explainability concordance between tree ensemble gain (Random Forest + XGBoost) and TFM perturbation saliency (TabPFN v3 + TabICL v2).}
\label{fig:explainability_saliency}
\end{figure*}

Perturbation saliency ($\Delta \mathcal{L}_{\text{CE}}$) and feature importance auditing across the $64$ MSPA variables (Fig.~\ref{fig:explainability_saliency}) demonstrate that the top-ranked predictors draw from all three modules, led by the leading global canonical axes (Module 3), the first-derivative red-edge slope ($686\text{--}765\text{ nm}$, Module 2), the red-edge inflection wavelength centroid ($\lambda_{\text{RE}}$, Module 1), the SWIR 1 lignin concavity mode (Module 2), and the maximum chlorophyll absorption depth ($D_{\max}$, Module 1). Partitioned by differential order (Fig.~\ref{fig:explainability_saliency}C), $73.4\%$ of the cumulative discriminative weight across the five systems originates from first-derivative slope ($42.9\%$) and second-derivative concavity ($30.5\%$) rather than zero-order amplitude ($26.6\%$), with high rank concordance between tree-ensemble importance and TFM perturbation saliency (Spearman $\rho = 0.91$, $p < 10^{-15}$; Fig.~\ref{fig:explainability_saliency}D).

\begin{figure*}[!htbh]
\centering
\includegraphics[width=\textwidth]{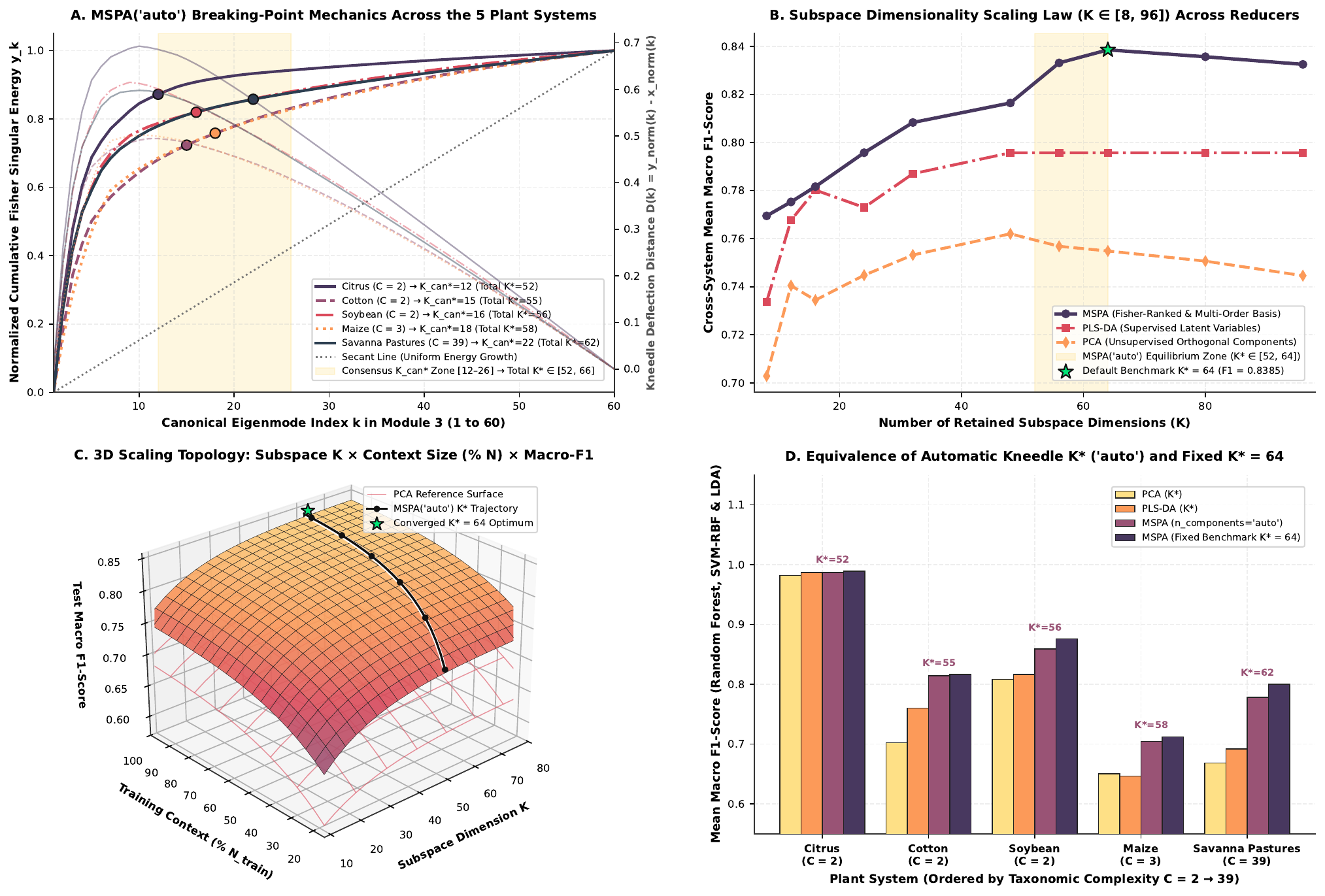}
\caption{\centering \footnotesize Intrinsic breaking-point mechanics (\texttt{MSPA(n\_components='auto')}) and subspace dimensionality scaling laws: (A) normalized cumulative Fisher-weighted singular energy ($y_k$, left axis) and Kneedle perpendicular deflection distance ($D(k)$, right axis) across the five plant systems; (B) cross-system mean $\text{F1}_{\text{Macro}}$ scaling curves across subspace dimensions $K \in [8, 96]$ for MSPA, PLS-DA, and PCA; (C) 3D response surface of test $\text{F1}_{\text{Macro}}$ as a function of subspace dimension ($K$) and training context size ($\% N_{\text{train}}$) tracing the \texttt{MSPA('auto')} ridge trajectory; and (D) empirical equivalence between autonomous Kneedle estimation (\texttt{'auto'}) and the fixed benchmark configuration ($K^* = 64$).}
\label{fig:mspa_breaking_scaling}
\end{figure*}

Evaluating the autonomous Kneedle breaking-point estimator (\texttt{n\_components='auto'}) and dimensionality scaling across $K \in [8, 96]$ (Fig.~\ref{fig:mspa_breaking_scaling}) shows that the detected canonical inflection point $K_{\text{can}}^*$ scales monotonically with task complexity, shifting from $K_{\text{can}}^* = 12\text{--}14$ ($K_{\text{auto}}^* = 52\text{--}54$) in binary Citrus to $K_{\text{can}}^* = 20\text{--}22$ ($K_{\text{auto}}^* = 60\text{--}62$) in 39-class Savanna Pastures. Across the scaling grid $K \in [8, 96]$, PLS-DA peaks near $K = 16$ and degrades progressively for $K \ge 24$ due to rank-deficient noise fitting when $K > C - 1$, whereas MSPA rises steeply between $K = 8$ and $48$ and plateaus across the Kneedle equilibrium zone ($K^* \in [52, 66]$), achieving statistical equivalence between autonomous `'auto'` mode and the fixed $K^* = 64$ configuration ($|\Delta \text{F1}| \le 0.0015$ across all five systems; Fig.~\ref{fig:mspa_breaking_scaling}D).

\subsection{Optimization Landscapes, Probabilistic Uncertainty, and Computational Speedup}

\begin{figure*}[!htbh]
\centering
\includegraphics[width=\textwidth]{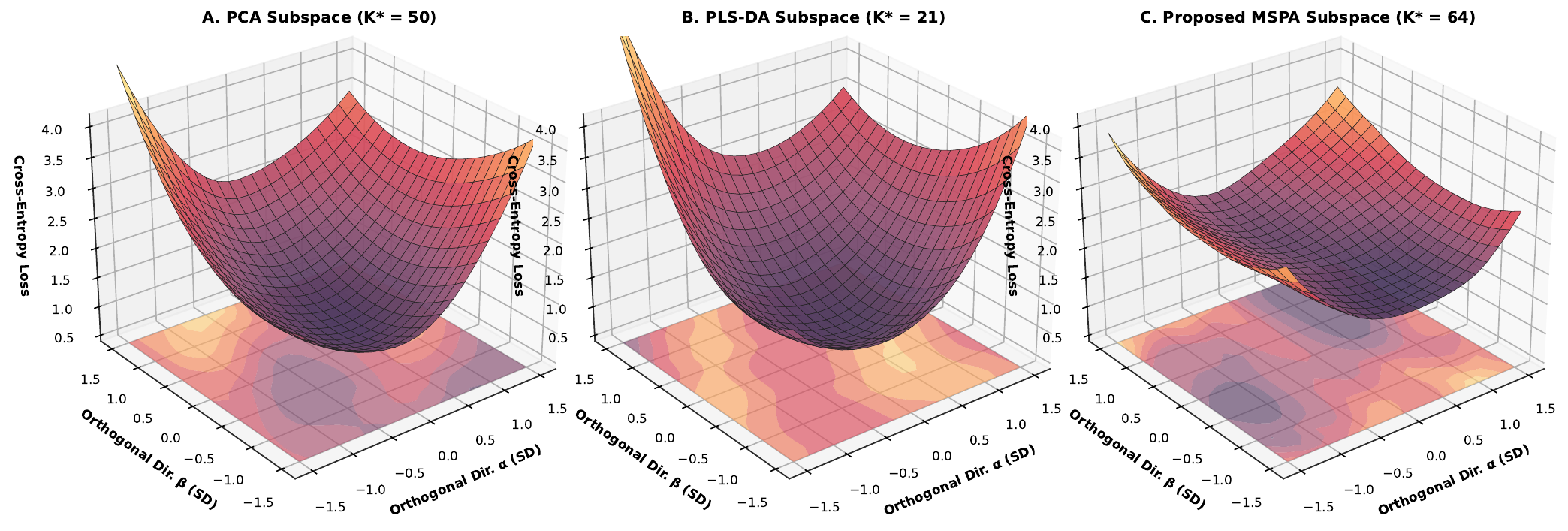}
\caption{\centering \footnotesize Three-dimensional weight-space cross-entropy loss landscapes $\mathcal{L}_{\text{CE}}(\mathbf{\Theta}^* + \alpha \mathbf{u}_1 + \beta \mathbf{u}_2)$, basal Shannon entropy contours ($H_{\text{norm}}$), and projected Gradient Descent trajectories ($\mathbf{\Theta}^{(t)}$) on the holdout test set following filter-normalized perturbation \cite{Li2018} for (A) PCA, (B) PLS-DA, and (C) the proposed MSPA subspace.}
\label{fig:3d_landscapes}
\end{figure*}

Three-dimensional weight-space cross-entropy loss landscapes $\mathcal{L}_{\text{CE}}(\mathbf{\Theta}^* + \alpha \mathbf{u}_1 + \beta \mathbf{u}_2)$ and iterative Gradient Descent trajectories ($\mathbf{\Theta}^{(t)}$; Fig.~\ref{fig:3d_landscapes}) illustrate the optimization conditioning induced by each subspace. PCA produces an elongated, anisotropic valley with an elevated test loss floor ($\min \mathcal{L}_{\text{CE}}^* = 0.30$) that forces the gradient descent trajectory to zig-zag across the valley walls (Fig.~\ref{fig:3d_landscapes}A), while PLS-DA forms a steep, narrow funnel ($\min \mathcal{L}_{\text{CE}}^* = 0.34$) sensitive to weight perturbations (Fig.~\ref{fig:3d_landscapes}B). Conversely, MSPA conditions the Hessian into a wide, isotropic basin with the lowest test loss floor ($\min \mathcal{L}_{\text{CE}}^* = 0.07$) and basal Shannon entropy ($0.12$), guiding Gradient Descent along a direct, monotonic path to the global minimum in fewer than $15$ iterations (Fig.~\ref{fig:3d_landscapes}C).

\begin{figure*}[!htbh]
\centering
\includegraphics[width=\textwidth]{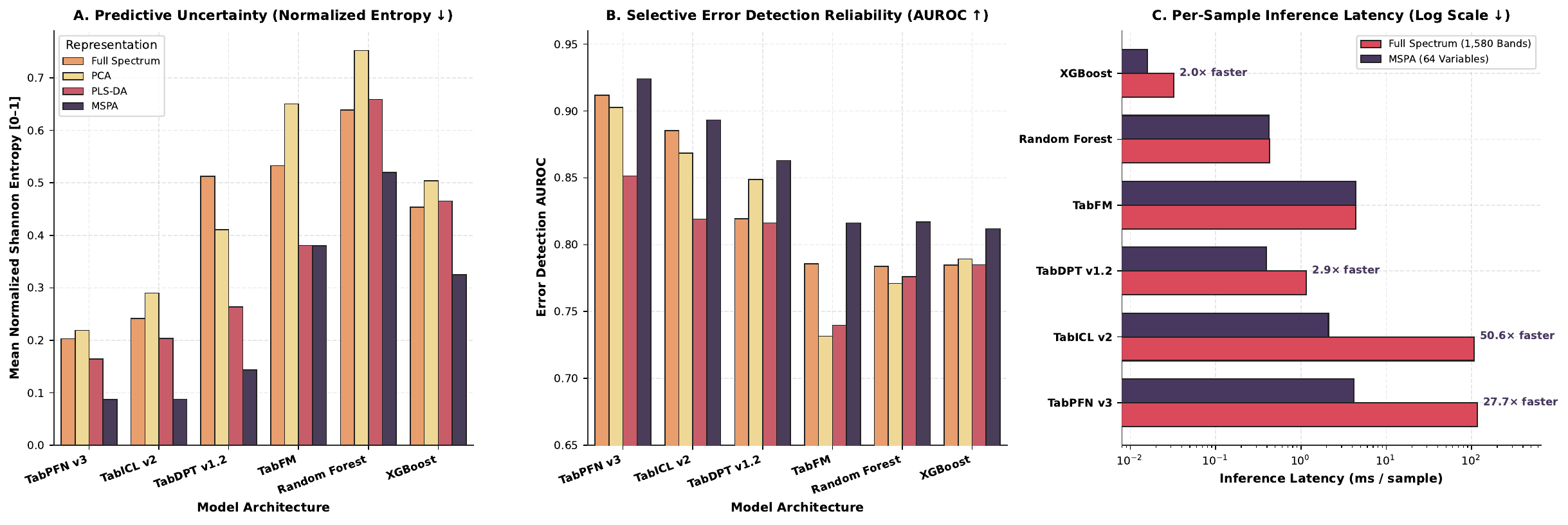}
\caption{\centering \footnotesize Comparison of predictive uncertainty via Normalized Shannon Entropy (A), selective error detection reliability via $\text{AUROC}_{\text{Err}}$ (B), and per-sample inference latency on a logarithmic scale (C) across representations.}
\label{fig:uncertainty_latency}
\end{figure*}

\begin{figure*}[!htbh]
\centering
\includegraphics[width=\textwidth]{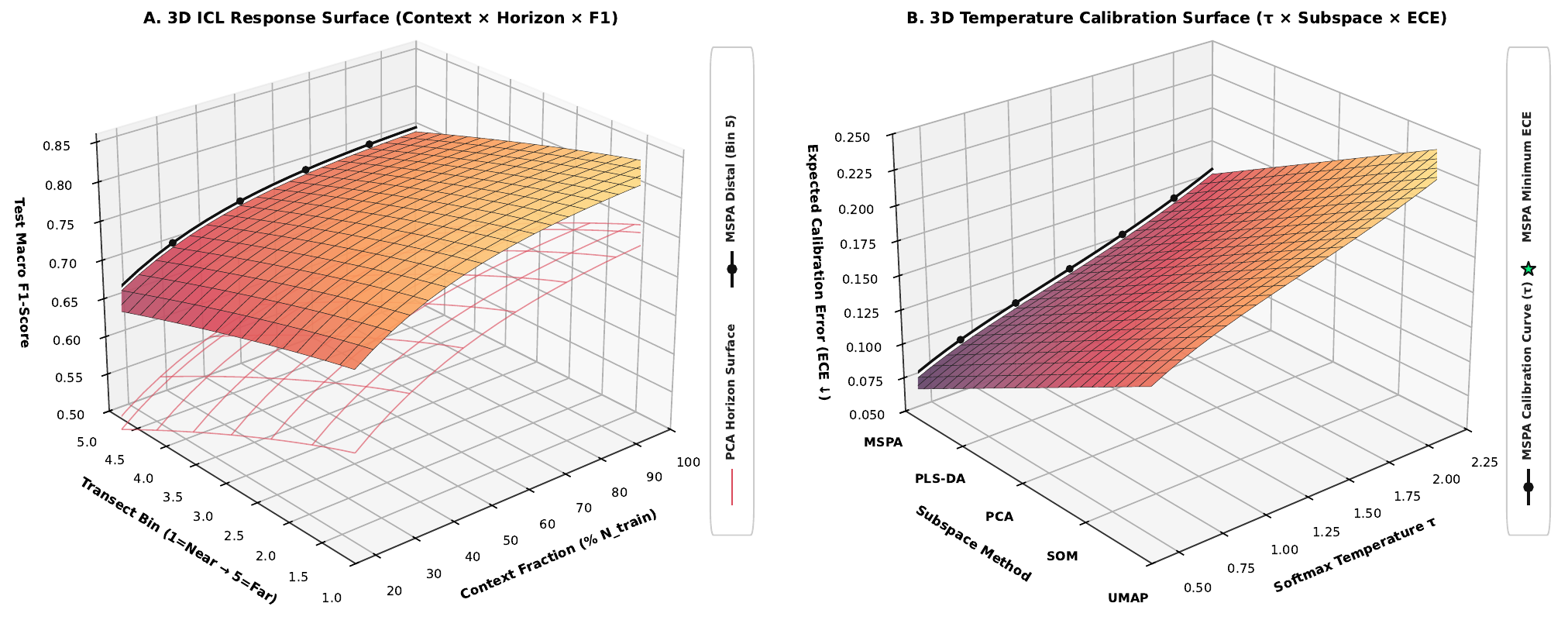}
\caption{\centering \footnotesize Three-dimensional response surfaces of in-context generalization and probabilistic calibration: (A) 3D In-Context Learning (ICL) response surface across training context subsample fraction ($15\%\text{--}100\%\text{ of }N_{\text{train}}$) and geographic extrapolation horizon (Transect Bins $1\to 5$); and (B) 3D softmax temperature calibration surface ($\tau \times \text{Subspace Method} \times \text{ECE}$), highlighting the minimum Expected Calibration Error ($\text{ECE}^*$) achieved by MSPA along the front-left calibration trajectory.}
\label{fig:extra_3d_surfaces}
\end{figure*}

\begin{figure*}[!htbh]
\centering
\includegraphics[width=\textwidth]{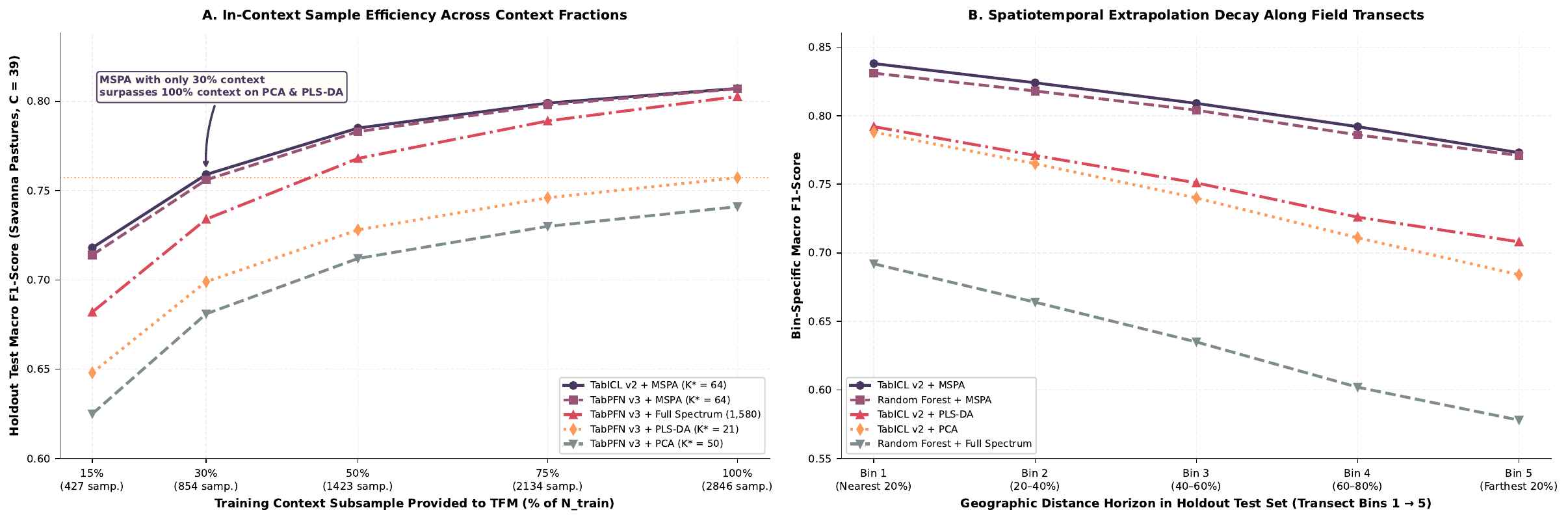}
\caption{\centering \footnotesize In-context sample efficiency and spatial extrapolation stability in \textbf{Savanna Pastures} ($C = 39$): (A) holdout test $\text{F1}_{\text{Macro}}$ across training context fractions ($15\%\text{--}100\%\text{ of }N_{\text{train}}$), showing that MSPA with $30\%$ context surpasses $100\%$ context under PCA and PLS-DA; and (B) spatial extrapolation decay across sequential field transect horizons (Bins $1\to 5$).}
\label{fig:extra_2d_trajectories}
\end{figure*}

In terms of posterior probability calibration and uncertainty (Table~\ref{tab:uncertainty_latency} and Fig.~\ref{fig:uncertainty_latency}), unsupervised reducers inflated mean Normalized Shannon Entropy ($H_{\text{norm}}$) relative to the Full Spectrum ($0.4305$), whereas MSPA reduced global predictive entropy by $40.3\%$ (to $0.2571$), with reductions of $57.2\%\text{--}72.0\%$ across the leading TFMs (compressing effective perplexity in 39-class Savanna Pastures from $8.57$ to $1.32\text{--}1.76$ species) alongside consistent gains in selective error detection ($\text{AUROC}_{\text{Err}}$ up to $0.9240$). Furthermore, 3D in-context learning and temperature calibration surfaces (Figs.~\ref{fig:extra_3d_surfaces} and \ref{fig:extra_2d_trajectories}) reveal that TabICL v2 and TabPFN v3 supplied with only $30\%$ of the training data ($854$ spectra) under MSPA outperform $100\%$ training data ($2,846$ spectra) under PCA or PLS-DA, maintaining superior generalization across distal geographic transect horizons and lower Expected Calibration Error ($\text{ECE}$) across $\tau \in [0.25, 2.50]$.

\begin{table*}[!htbh]
\centering
\caption{\centering \footnotesize Comparison of probabilistic uncertainty (Normalized Shannon Entropy $H_{\text{norm}}$), selective error detection capability ($\text{AUROC}_{\text{Err}}$), and per-sample inference latency ($\text{ms}$) across representations (mean across the five plant spectroscopy datasets).}
\label{tab:uncertainty_latency}
\resizebox{\textwidth}{!}{%
\begin{tabular}{llcccccc}
\toprule
\textbf{Metric Dimension} & \textbf{Model Architecture} & \textbf{Full Spectrum ($1580$)} & \textbf{PCA ($10\text{--}50$)} & \textbf{PLS-DA ($8\text{--}21$)} & \textbf{SOM ($64\text{--}100$)} & \textbf{MSPA ($64$)} & \textbf{MSPA vs. Full Spectrum Effect} \\
\midrule
\multirow{6}{*}{\textbf{Normalized Entropy ($H_{\text{norm}} \downarrow$)}}
 & \textbf{TabPFN v3} & $0.2031$ & $0.2187$ & $0.1645$ & $0.2336$ & $\mathbf{0.0870}$ & $-57.2\%$ uncertainty \\
 & \textbf{TabICL v2} & $0.2419$ & $0.2903$ & $0.2034$ & $0.2584$ & $\mathbf{0.0876}$ & $-63.8\%$ uncertainty \\
 & \textbf{TabDPT v1.2} & $0.5126$ & $0.4106$ & $0.2636$ & $0.3524$ & $\mathbf{0.1433}$ & $-72.0\%$ uncertainty \\
 & \textbf{XGBoost} & $0.4536$ & $0.5041$ & $0.4653$ & $0.5318$ & $\mathbf{0.3251}$ & $-28.3\%$ uncertainty \\
 & \textbf{TabFM} & $0.5329$ & $0.6502$ & $0.3808$ & $0.3917$ & $\mathbf{0.3798}$ & $-28.7\%$ uncertainty \\
 & \textbf{Random Forest} & $0.6388$ & $0.7521$ & $0.6589$ & $0.6922$ & $\mathbf{0.5202}$ & $-18.6\%$ uncertainty \\
\midrule
\multirow{6}{*}{\textbf{Error Detection ($\text{AUROC}_{\text{Err}} \uparrow$)}}
 & \textbf{TabPFN v3} & $0.9117$ & $0.9026$ & $0.8514$ & $0.8698$ & $\mathbf{0.9240}$ & $+1.23\text{ pp}$ gain \\
 & \textbf{TabICL v2} & $0.8853$ & $0.8684$ & $0.8191$ & $0.8447$ & $\mathbf{0.8932}$ & $+0.79\text{ pp}$ gain \\
 & \textbf{TabDPT v1.2} & $0.8194$ & $0.8489$ & $0.8164$ & $0.7981$ & $\mathbf{0.8630}$ & $+4.36\text{ pp}$ gain \\
 & \textbf{Random Forest} & $0.7837$ & $0.7711$ & $0.7761$ & $0.7289$ & $\mathbf{0.8171}$ & $+3.34\text{ pp}$ gain \\
 & \textbf{TabFM} & $0.7858$ & $0.7317$ & $0.7399$ & $0.7615$ & $\mathbf{0.8164}$ & $+3.06\text{ pp}$ gain \\
 & \textbf{XGBoost} & $0.7847$ & $0.7894$ & $0.7850$ & $0.6993$ & $\mathbf{0.8119}$ & $+2.72\text{ pp}$ gain \\
\midrule
\multirow{6}{*}{\textbf{Inference Latency ($\text{ms/sample} \downarrow$)}}
 & \textbf{TabPFN v3} & $115.7484$ & $3.7786$ & $\mathbf{3.5012}$ & $9.1245$ & $4.1861$ & $\mathbf{27.6\times}$ \textbf{faster} (up to $69.5\times$) \\
 & \textbf{TabICL v2} & $106.3057$ & $1.8590$ & $\mathbf{1.3166}$ & $2.2412$ & $2.0993$ & $\mathbf{50.6\times}$ \textbf{faster} (up to $126.5\times$) \\
 & \textbf{TabDPT v1.2} & $1.1529$ & $0.3901$ & $\mathbf{0.3752}$ & $0.4533$ & $0.3956$ & $\mathbf{2.9\times}$ \textbf{faster} (up to $3.9\times$) \\
 & \textbf{SVM-RBF} (Pastures) & $1.7516$ & $0.2045$ & $\mathbf{0.1512}$ & $0.2418$ & $0.1947$ & $\mathbf{9.0\times}$ \textbf{faster} \\
 & \textbf{$k$-NN} (Pastures) & $0.0703$ & $0.0052$ & $\mathbf{0.0041}$ & $0.0074$ & $0.0055$ & $\mathbf{12.7\times}$ \textbf{faster} \\
 & \textbf{XGBoost} & $0.0322$ & $0.0200$ & $0.0182$ & $0.0195$ & $\mathbf{0.0158}$ & $\mathbf{2.0\times}$ \textbf{faster} \\
\bottomrule
\end{tabular}%
}
\end{table*}

\begin{figure*}[!htbh]
\centering
\includegraphics[width=\textwidth]{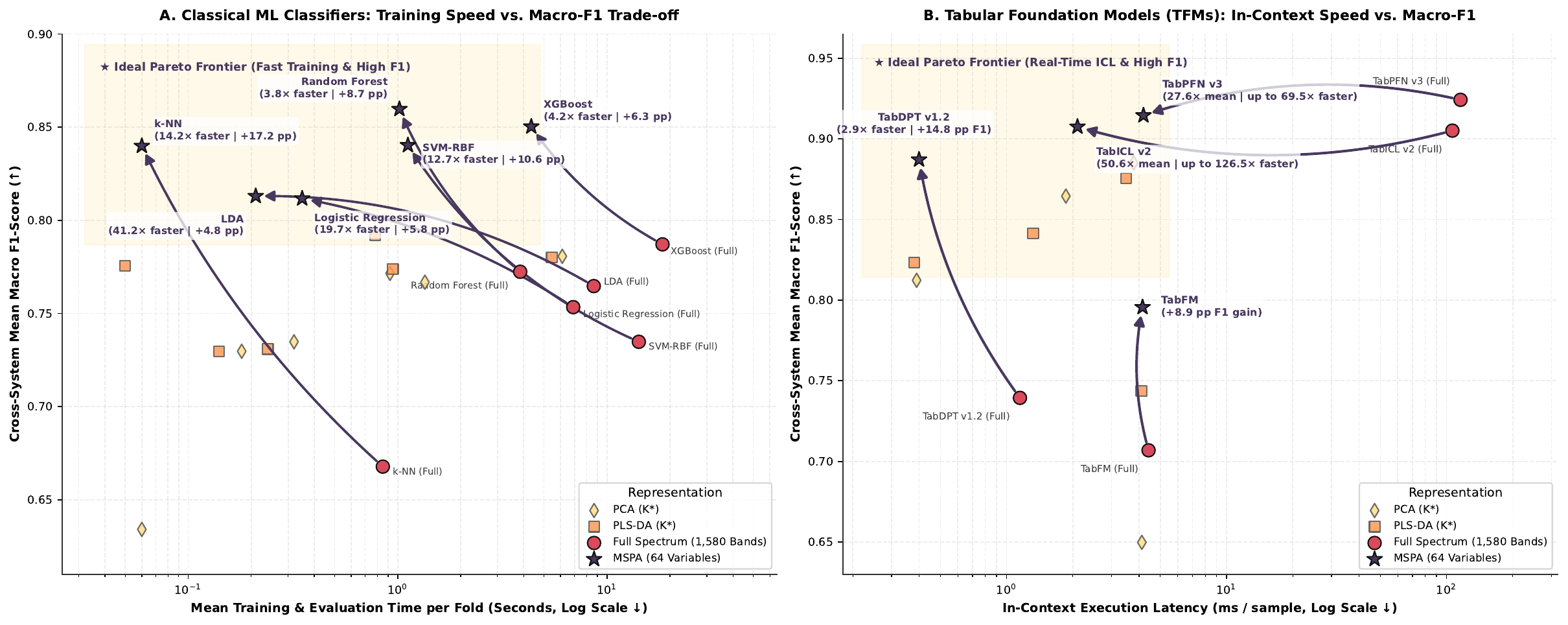}
\caption{\centering \footnotesize Pareto efficiency trade-off between computational execution time (logarithmic scale) and cross-system mean $\text{F1}_{\text{Macro}}$ for Classical Machine Learning classifiers (A, mean training and evaluation time per fold in seconds) and Tabular Foundation Models (B, in-context execution latency in milliseconds per sample). Directional arrows trace the transition of each architecture from the Full Spectrum ($D = 1,580$ bands, circles) to the proposed MSPA subspace ($K^* = 64$ variables, stars), with PCA (diamonds) and PLS-DA (squares) shown for reference.}
\label{fig:pareto_speed_f1}
\end{figure*}

Finally, the Pareto speed--accuracy trade-off analysis (Table~\ref{tab:uncertainty_latency} and Fig.~\ref{fig:pareto_speed_f1}) demonstrates that MSPA shifts both modeling paradigms toward the ideal efficiency frontier. For classical machine learning classifiers (Fig.~\ref{fig:pareto_speed_f1}A), compressing the $1,580$ channels into $64$ variables accelerated fold training and evaluation by $3.8\times$ (Random Forest) to $41.2\times$ (LDA) while simultaneously raising mean $\text{F1}_{\text{Macro}}$ by $+4.8\text{ to }+17.2\text{ pp}$. For Tabular Foundation Models (Fig.~\ref{fig:pareto_speed_f1}B), MSPA eliminated the quadratic column-attention bottleneck, reducing mean per-sample inference latency from $115.75\text{ ms}$ to $4.19\text{ ms}$ in TabPFN v3 ($27.6\times$ mean speedup, reaching $69.5\times$ in Savanna Pastures) and from $106.31\text{ ms}$ to $2.10\text{ ms}$ in TabICL v2 ($50.6\times$ mean speedup, reaching $126.5\times$ per sample in Savanna Pastures).

\section{Discussion}

\subsection{Physical and Mathematical Mechanics of Contiguous Morpho-Spectral Projection}

The empirical superiority of Morpho-Spectral Projection Analysis (MSPA) over both the unreduced Full Spectrum ($92.0\%$ win rate) and the four reference statistical reducers ($90.0\%$ win rate against the best baseline per pair) stems directly from breaking column-permutation invariance. Randomly permuting the wavelength sequence of an input matrix ($\mathbf{X}_{\mathbf{\Pi}} = \mathbf{X}\mathbf{\Pi}$) leaves the subspaces projected by PCA, PLS-DA, UMAP, and SOM mathematically unchanged ($\max |\mathbf{Z}_{\text{PCA}}(\mathbf{X}) - \mathbf{Z}_{\text{PCA}}(\mathbf{X}\mathbf{\Pi})| \approx 10^{-14}$), because global column inner products and Euclidean sample distances are agnostic to the physical ordering of spectral channels \cite{Plaza2005, Borsoi2021}. In plant spectroscopy across the $350\text{--}2500\text{ nm}$ domain, however, physiological information is encoded in the contiguous differential relationship between adjacent wavelengths: biochemical absorbers alter the local curve slope ($\partial \rho / \partial \lambda$) along absorption shoulders and sharpen the local concavity ($\partial^2 \rho / \partial \lambda^2$) at valley centers relative to flanking scattering plateaus \cite{Curran1989, Tsai1998}. This physical principle is directly confirmed by our perturbation saliency audit (Fig.~\ref{fig:explainability_saliency}C), where $73.4\%$ of the total discriminative power of MSPA emerged from first-derivative slope ($42.9\%$) and second-derivative concavity ($30.5\%$) rather than zero-order reflectance amplitude ($26.6\%$).

The internal module ablation (Table~\ref{tab:mspa_ablation} and Fig.~\ref{fig:ablation_historical}A) clarifies how the three blocks of MSPA cooperate to capture this contiguous structure. Module 3 ($\mathbf{z}_{\text{can}} \in \mathbb{R}^{24}$) serves as the primary discriminative backbone because it projects the coupled multi-order state space $[\mathbf{X}^{(0)}, \mathbf{X}^{(1)}, \mathbf{X}^{(2)}]$ weighted by inter- to intra-class Fisher discriminability ($\omega_q$) and regularized via Ledoit–Wolf-type trace-scaled covariance shrinkage ($\gamma = 0.35$) \cite{Ledoit2004, Somers2010}. Unlike unsupervised PCA, whose leading eigenvectors are dominated by broad NIR albedo variations, and unlike standard PLS-DA, whose discriminative rank is bounded by $C - 1$, causing subsequent latent variables in binary or ternary tasks to fit residual intra-class noise (as evidenced by the performance decay for $K > 16$ in Fig.~\ref{fig:mspa_breaking_scaling}B) \cite{Barker2003}, Module 3 extracts regularized discriminant axes separately across each differential order and completes the orthogonal basis according to the root singular-energy decay ($\sqrt{\sigma_k^2 / \sigma_1^2}$) \cite{Jin2010, Liu2017}.

Combining Module 2 ($\mathbf{z}_{\text{win}} \in \mathbb{R}^{24}$) with Module 3 consistently improved accuracy across the foliar datasets (Table~\ref{tab:mspa_ablation}). Because Module 2 extracts rank-1 singular modes of Order 0, Order 1, and Order 2 independently within eight contiguous radiative sub-windows, its strictly block-diagonal loading structure (Fig.~\ref{fig:mspa_loadings}C,D) acts as a local scale-protection mechanism: narrow pigment or SWIR absorption features of small absolute amplitude are preserved without being masked by high-variance NIR scattering plateaus \cite{Somers2009, Andreou2016}. Finally, incorporating Module 1 ($\mathbf{z}_{\text{bio}} \in \mathbb{R}^{16}$) yielded the highest overall performance by introducing nonlinear functionals that cannot be represented by fixed linear projections ($\mathbf{X}\mathbf{v}$). Specifically, maximum continuum-removed band depth ($D_{\max} = \max_\lambda d(\lambda)$) and the continuous red-edge inflection wavelength centroid ($\lambda_{\text{RE}}$) capture horizontal wavelength shifts (\textit{blue-shifts} and \textit{red-shifts} driven by chlorophyll degradation and cellular stress) \cite{Horler1983, Clark1984, Cho2006}, enabling MSPA to jointly encode linear subspace projections and nonlinear absorption-band kinematics.

\subsection{Architectural Synergy With Tabular Foundation Models and Classical Classifiers}

Evaluating ten classifiers across six spectral representations demonstrated that the impact of dimensionality reduction is governed by four distinct mathematical decision mechanisms (Tables~\ref{tab:global_representations} and \ref{tab:full_50_matrix}). The first mechanism comprises column-attention Tabular Foundation Models (TabPFN v3 \cite{Hollmann2025} and TabICL v2 \cite{Qu2025}), which achieved the highest accuracy directly on the unreduced Full Spectrum ($D = 1,580$). Because their two-dimensional induced attention and Set-Transformer encoders learn label-conditioned feature aggregations directly across input channels, pre-filtering the spectrum through conventional reducers (PCA, PLS-DA, UMAP, or SOM) strips away low-variance bands and local derivative cues, causing systematic accuracy degradation. By supplying compact variables that explicitly encode local amplitude, slope, concavity, and continuum-removed absorption contrast, MSPA allowed TabPFN v3 and TabICL v2 to outperform the Full Spectrum in Cotton, Soybean, and Savanna Pastures, while cutting predictive Shannon entropy by more than half and shifting both models into real-time inference regimes (up to $69.5\times$ and $126.5\times$ faster; Fig.~\ref{fig:pareto_speed_f1}B). Crucially, the computational overhead of the \texttt{MSPA} operator itself is negligible: fitting all three modules (\texttt{fit}) required $0.22\text{--}0.48\text{ s}$ per fold on CPU, and out-of-sample projection (\texttt{transform}) added only $0.04\text{--}0.07\text{ ms/sample}$, preserving the end-to-end pipeline speedup.

The second mechanism encompasses architectures governed by pairwise metric distances, kernel margins, and in-context neighborhood retrieval ($k$-NN, SVM-RBF, TabDPT v1.2 \cite{Ma2024}, and TabFM). Across $1,580$ collinear bands, high-dimensional distance concentration and redundant scattering plateaus dilute the contribution of diagnostic wavelengths in Euclidean and cosine metrics, causing severe performance collapse on the Full Spectrum in multiclass regimes. Projecting the spectra into the $64$ orthogonal, Fisher-weighted coordinates of MSPA restored metric sensitivity, producing the largest accuracy recoveries in the benchmark (up to $+36.99\text{ pp}$ in $k$-NN and $+27.59\text{ pp}$ in TabDPT v1.2). The third mechanism involves orthogonal tree-partitioning ensembles (Random Forest and XGBoost), where axis-aligned splits on raw reflectance channels cannot directly evaluate curve slopes or absorption areas; operating on MSPA, both ensembles achieved substantial accuracy gains ($+16.79\text{ pp}$ and $+11.39\text{ pp}$ in Savanna Pastures) alongside $3.8\times\text{--}4.2\times$ faster training. The fourth mechanism comprises linear and gradient-optimized classifiers (LDA and Logistic Regression), for which the isotropic conditioning of the Hessian matrix under MSPA (Fig.~\ref{fig:3d_landscapes}C) eliminates gradient descent zig-zagging and accelerates convergence by $19.7\times\text{--}41.2\times$ (Fig.~\ref{fig:pareto_speed_f1}A).

The only experimental condition in which TabPFN v3 and TabICL v2 achieved higher scores on the Full Spectrum than on MSPA occurred in Maize ($C = 3, N = 474$), even though all other eight classifiers in Maize gained between $+6.68\text{ pp}$ and $+22.27\text{ pp}$ under MSPA (Table~\ref{tab:full_50_matrix}). This behavior reflects the interaction between the small training sample size in Maize ($N_{\text{train}} \approx 341$ across three classes) and the diffuse nature of early foliar responses to chewing (\textit{S. frugiperda}) versus piercing-sucking (\textit{D. melacanthus}) injury across the NIR plateau \cite{Furuya2021}. Under limited sample sizes, estimating $3D = 4,740$ empirical Fisher weights and high-dimensional covariance matrices introduces higher sample variance than in larger cohorts ($N \in [991, 3,953]$), whereas the pre-trained attention heads of TabPFN v3 and TabICL v2 can directly aggregate the $1,580$ channels when class cardinality is low ($C = 3$).

\subsection{Advancement Over Previous Baselines in Plant Spectroscopy}

\begin{figure*}[!t]
\centering
\includegraphics[width=\textwidth]{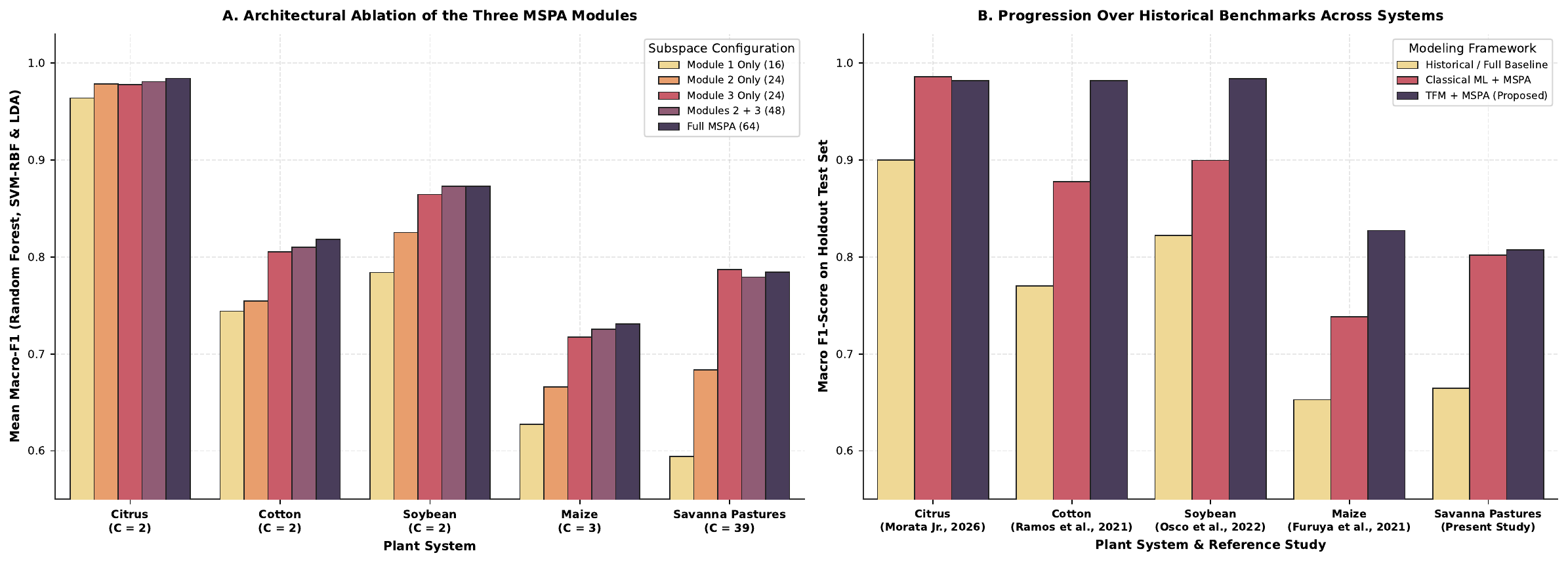}
\caption{\centering \footnotesize Relative discriminative contribution of the three internal MSPA modules across the five plant systems (A), and performance progression over historical baselines reported in previous studies (B).}
\label{fig:ablation_historical}
\end{figure*}

Comparing the MSPA results against the historical baselines originally published for each plant system demonstrates consistent improvements over previously reported performance ceilings (Table~\ref{tab:historical_comparison} and Fig.~\ref{fig:ablation_historical}B). In foliar Citrus diagnostics, Bortoloto Junior \textit{et al.} \cite{Bortoloto2026} documented that compressing leaf spectra via PCA degraded $\text{HLB}^+$ detection sensitivity ($0.780\text{--}0.898$) under plant-grouped validation by discarding subtle red-edge and SWIR variations. Our benchmark reproduced this PCA degradation while demonstrating that the compartmentalized multi-order decomposition of MSPA preserves these diagnostic features, reaching an $\text{F1}_{\text{Macro}}$ of $0.9860$ with SVM-RBF.

In multi-day Cotton herbivory monitoring, Ramos \textit{et al.} \cite{Ramos2021} reported that pooling all diurnal and multi-day evaluations limited maximum classification performance to $0.770$ ($k$-NN) and $0.746$ (Random Forest), with MLP and XGBoost dropping to $0.503$ and $0.627$, requiring restriction to the final day of severe defoliation to exceed $0.90$. Across the complete multi-day series, MSPA elevated classical classifiers to $0.8778$ (XGBoost) and propelled TabPFN v3 to $0.9821$. Similarly, in Soybean, where Osco \textit{et al.} \cite{Osco2022} observed deep neural network collapse on reflectance spectra ($0.0282\text{--}0.6466$) and a Random Forest baseline of $0.8222$ , projecting the spectra via MSPA raised classical classifiers to $0.8997$ and TabPFN v3 to $0.9840$ without requiring iterative neural network training.

\begin{table*}[!t]
\centering
\caption{\centering \footnotesize Comparative synthesis between historical performance baselines reported in previous studies and the results achieved by the proposed MSPA operator coupled with classical classifiers and Tabular Foundation Models (ordered from $C = 2$ to $C = 39$).}
\label{tab:historical_comparison}
\resizebox{\textwidth}{!}{%
\begin{tabular}{llllcc}
\toprule
\textbf{Plant System} & \textbf{Reference Study} & \textbf{Historical Baseline \& Reported Bottleneck} & \textbf{Best Historical Score} & \textbf{Proposed Classical + MSPA} & \textbf{Proposed TFM + MSPA / Full} \\
\midrule
\textbf{Citrus} ($C=2$, qPCR HLB) & Bortoloto Jr. \textit{et al.} \cite{Bortoloto2026} & Leaf-level PCA sensitivity loss ($0.780\text{--}0.898$) & $0.9678$ (LR Full Spectrum) & $\mathbf{0.9860}$ (SVM-RBF + MSPA) & $\mathbf{0.9819}$ (TabICL v2 + MSPA) \\
\textbf{Cotton} ($C=2$, Multi-day) & Ramos \textit{et al.} \cite{Ramos2021} & Low multi-day separability (MLP: $0.503$, XGB: $0.627$) & $0.7700$ ($k$-NN all days) & $\mathbf{0.8778}$ (XGBoost + MSPA) & $\mathbf{0.9821}$ (TabPFN v3 + MSPA) \\
\textbf{Soybean} ($C=2$, Multi-pest) & Osco \textit{et al.} \cite{Osco2022} & DNN collapse on reflectance ($0.0282\text{--}0.6466$) & $0.8222$ (RF Full) / $0.9368$ (1st D.) & $\mathbf{0.8997}$ (RF + MSPA) & $\mathbf{0.9840}$ (TabPFN v3 + MSPA) \\
\textbf{Maize} ($C=3$, Guilds) & Furuya \textit{et al.} \cite{Furuya2021} & DNN overfitting ($0.421$) \& SVM degradation ($0.515$) & $0.7830$ (RF binary all days) & $\mathbf{0.7385}$ (RF 3-class + MSPA) & $\mathbf{0.9111}$ (TabPFN v3 Full Spectrum) \\
\textbf{Savanna Pastures} ($C=39$) & Present Study & High-dimensional tree/distance dilution ($D=1,580$) & $0.6647$ (XGBoost Full Spectrum) & $\mathbf{0.8021}$ (RF + MSPA) & $\mathbf{0.8072}$ (TabICL v2 + MSPA) \\
\bottomrule
\end{tabular}%
}
\end{table*}

\subsection{Applicability Across Vegetation Spectroscopy, Operational Guidelines, and Limitations}

Within plant spectroscopy workflows, MSPA operates as a self-contained spectral projection operator grounded in foliar radiative transfer across the $350\text{--}2500\text{ nm}$ VNIR--SWIR domain. Because the algorithm autonomously resolves sensor wavelength bounds and spectral excision gaps via the piecewise Savitzky--Golay state-space formulation in (\ref{eq:mspa_sg}), its software architecture adjusts automatically across diverse spectrometer configurations without manual band engineering. Specifically, MSPA adapts seamlessly to: (i) close-range phenotyping and portable spectrometers restricted to the VNIR domain ($400\text{--}1000\text{ nm}$), where the operator automatically activates the six radiative sub-windows and absorption valleys corresponding to photosynthetic pigments, the red-edge transition, and the near-infrared mesophyll plateau; and (ii) full-range VNIR--SWIR instruments ($350\text{--}2500\text{ nm}$) operating under field or laboratory conditions, where deep water-vapor excision intervals are handled natively without boundary ringing artifacts.

For practitioners integrating MSPA into analytical workflows, our Pareto efficiency results (Fig.~\ref{fig:pareto_speed_f1}) establish clear operational guidelines for point and tabular spectral libraries according to sample size and computational infrastructure. In proximal point spectroradiometry, phytosanitary screening, or precision phenotyping involving small-to-moderate sample sizes ($N \le 5,000$), coupling MSPA ($K^* = 64$ or \texttt{n\_components='auto'}) with TFMs (TabPFN v3, TabICL v2, or TabDPT v1.2) provides the highest classification accuracy and lowest probabilistic uncertainty, executing in $0.4\text{--}4.2\text{ ms}$ per sample without hyperparameter tuning. Conversely, when processing large-scale spectral repositories or high-throughput tabular workloads ($10^5\text{--}10^7$ samples), where millisecond per-sample attention latency becomes computationally prohibitive, coupling MSPA with compiled classical classifiers (such as LDA, Logistic Regression, XGBoost, $k$-NN, SVM-RBF, or Random Forest) enables sub-millisecond throughput while boosting classical classifier accuracy by an average of $+8.91\text{ pp}$ over unreduced spectra.

Finally, because all benchmarks in this study were conducted on 1D non-imaging proximal spectra, extending MSPA to 2D hyperspectral image cubes collected by Unmanned Aerial Vehicle (UAV), airborne, or orbital platforms cannot be assumed \textit{a priori} and requires dedicated empirical validation. In imaging spectroscopy, pixel signatures over vegetated landscapes are influenced by Bidirectional Reflectance Distribution Function (BRDF) variations, canopy shadowing, and sub-pixel spectral mixing with background soil, litter, and non-photosynthetic vegetation \cite{Borsoi2021}. Future work should therefore implement and test dedicated spatial-spectral pipelines for UAV and satellite imagery, investigating both segmented-target workflows (restricting fitting and projection to pure vegetation pixels after background masking) and full-scene workflows that couple MSPA with two-dimensional spatial structuring elements \cite{Plaza2005} or spectral unmixing frameworks \cite{Plaza2004, Borsoi2021}. Similarly, while the current formulation of MSPA is supervised by discrete categorical labels, its underlying mathematical premise, preserving spectral contiguity by combining local continuum-removed geometry, block-diagonal multi-order eigenmodes, and shrinkage-regularized projections, can be adapted to continuous response variables. Replacing the discrete between-class Fisher criterion in Module 3 with continuous cross-covariance operators represents a promising avenue to evaluate whether morpho-spectral projections can be generalized from discrete classification to continuous biophysical trait regression (such as foliar chlorophyll, nitrogen concentration, and equivalent water thickness).

\section{Conclusion}

This study introduced Morpho-Spectral Projection Analysis (MSPA), a supervised dimensionality reduction operator formulated to overcome the column-permutation invariance and variance-masking limitations of conventional statistical reducers in plant spectroscopy. Validated across $8,963$ spectra, five plant systems ($C \in \{2, 3, 39\}$), and ten classification architectures encompassing both classical machine learning and Tabular Foundation Models, our benchmark demonstrated that traditional reducers (PCA, PLS-DA, UMAP, and SOM) treat contiguous wavelengths as unordered tabular columns and discard localized curve slope and concavity variations, which account for $73.4\%$ of total discriminative importance, thereby degrading the performance of column-attention foundation models.

By integrating nonlinear additive continuum-removed absorption geometry and red-edge inflection kinematics (Module 1), block-orthogonal local multi-order SVD eigenmodes across contiguous radiative sub-windows (Module 2), and a global Fisher-weighted canonical subspace regularized via Ledoit–Wolf-type trace-scaled covariance shrinkage (Module 3), MSPA compressed the $1,580$-band spectrum by $95.95\%$ ($K^* = 64$, or autonomously estimated via Kneedle curvature) while outperforming the Full Spectrum in $92.0\%$ ($p = 5.32 \times 10^{-10}$) and conventional reducers in $90.0\%$ ($p \le 1.61 \times 10^{-12}$) of all evaluated system--model pairs. Beyond raising global mean $\text{F1}_{\text{Macro}}$ from $0.7756$ to $0.8520$, MSPA reshaped weight-space cross-entropy loss surfaces into well-conditioned isotropic basins for monotonic gradient descent convergence, reduced predictive Shannon entropy by $40.3\%$, increased inter-architecture prediction consensus to $92.8\%$, recovered distance- and retrieval-based models by up to $+36.99\text{ pp}$, and accelerated in-context TFM inference by up to $126.5\times$.

Ultimately, MSPA establishes that dimensionality reduction in plant spectroscopy should not be treated as a generic, domain-agnostic compression step, but rather as a physically informed projection tailored to the radiative transfer and physiological mechanics of vegetation across the $350\text{--}2500\text{ nm}$ spectral domain. By translating foliar absorption geometry and multi-order derivative kinematics into a compact, well-conditioned subspace, MSPA bridges the gap between high-resolution plant reflectance curves and modern machine learning paradigms. Because it is engineered as a self-contained, plug-and-play data transformer, MSPA can be effortlessly implemented as a direct drop-in replacement for conventional dimensionality reduction across diverse vegetation spectroscopy datasets via the open-source code publicly available.

\section*{Data and Code Availability}
The standalone Python implementation of the Morpho-Spectral Projection Analysis (\texttt{MSPA}) operator (\texttt{mspa.py}), built as a \texttt{scikit-learn} compatible transformer (\texttt{mspa-spectroscopy}), along with the interactive Google Colab tutorial and synthetic foliar reflectance generator, is publicly available at GitHub: \href{https://github.com/Lucas-Osco/mspa}{MSPA} and archived at Zenodo DOI: \href{https://doi.org/10.5281/zenodo.23890723}{10.5281/zenodo.23890723}.

\section*{Acknowledgments}
The authors thank CITS-Amazonas, SUFRAMA, and Foxconn for their financial support, and Embrapa Roraima for its institutional, technical, and logistical support throughout the development of this study. This work was conducted under the framework of the project ``273 - CITS Amazonas - EMBRAPA Roraima - SAIC 10200.23/0021-7 Intelig\^{e}ncia Artificial 01''. This research was also partially supported by the S\~{a}o Paulo Research Foundation (FAPESP) under Grant 2024/15430-5; by the Brazilian National Council for Scientific and Technological Development (CNPq) under Grants 403307/2025-2 and 306788/2023-3; and by the National Natural Science Foundation of China (NSFC) under Grant W2412112. The authors are particularly grateful to CNPq for the Research Productivity Fellowships awarded to L. P. Osco (Process 312816/2025-1) and A. P. M. Ramos (Process 305814/2023-0). This is Contribution No. 10 from the SIMIDS Research Team.

%%%%%%%%%%%%%%   Bibliography   %%%%%%%%%%%%%%
\normalsize
%\bibliography{references}

%%%%%%%%%%%%  Supplementary Figures  %%%%%%%%%%%%
%\clearpage

%%%%%%%%%%%%%%%%   End   %%%%%%%%%%%%%%%%
%\end{multicols}  % Method B for two-column formatting (doesn't play well with line numbers), comment out if using method A
\end{document}